\documentclass[10pt,journal,compsoc]{IEEEtran}
  \usepackage{cite}

\usepackage{enumitem}

\ifCLASSINFOpdf
\else
\fi
\usepackage{hyperref}

\usepackage{url}

\usepackage[english]{babel}
\usepackage[utf8]{inputenc}
\usepackage{amsmath}
\usepackage{graphicx}

\usepackage[colorinlistoftodos]{todonotes}
\usepackage{comment}
\usepackage{tikz}
\usepackage{subfig}

\def\checkmark{\tikz\fill[scale=0.35](0,.35) -- (.25,0) -- (1,.7) -- (.25,.15) -- cycle;}

\begin{document}
%
\title{Multimodal Machine Learning:\\
 A Survey and Taxonomy}
%
%
%
%

\author{Tadas~Baltru\v{s}aitis,
        Chaitanya~Ahuja,
        and~Louis-Philippe Morency
\IEEEcompsocitemizethanks{\IEEEcompsocthanksitem T. Baltru\v{s}aitis, C. Ahuja and L-P. Morency are with the Language Technologies Institute, at Carnegie Mellon University, Pittsburgh, Pennsylvania \protect\\
E-mail: tbaltrus, cahuja, morency@cs.cmu.edu}
}

%
%

\markboth{}%
{Shell \MakeLowercase{\textit{et al.}}: Bare Demo of IEEEtran.cls for Computer Society Journals}
%



\IEEEtitleabstractindextext{%
\begin{abstract}
Our experience of the world is multimodal  - we see objects, hear sounds, feel texture, smell odors, and taste flavors. 
\textit{Modality} refers to the way in which something happens or is experienced and a research problem is characterized as \textit{multimodal} when it includes multiple such modalities.
In order for Artificial Intelligence to make progress in understanding the world around us, it needs to be able to interpret such multimodal signals together. 
\textit{Multimodal machine learning} aims to build models that can process and relate information from multiple modalities. 
It is a vibrant multi-disciplinary field of increasing importance and with extraordinary potential.
Instead of focusing on specific multimodal applications, this paper surveys the recent advances in multimodal machine learning itself and presents them in a common taxonomy. 
We go beyond the typical early and late fusion categorization and identify broader challenges that are faced by multimodal machine learning, namely: representation, translation, alignment, fusion, and co-learning.
This new taxonomy will enable researchers to better understand the state of the field and identify directions for future research.

\end{abstract}

}

\maketitle

\IEEEdisplaynontitleabstractindextext

%
\IEEEpeerreviewmaketitle

\IEEEraisesectionheading{\section{Introduction}\label{sec:introduction}}

%
%
%
%


\label{sec:intro}

\IEEEPARstart{T}{he} world surrounding us involves multiple modalities --- we see objects, hear sounds, feel texture, smell odors, and so on. 
In general terms, a \textit{modality} refers to the way in which something happens or is experienced. 
Most people associate the word modality with the \textit{sensory modalities} which represent our primary channels of communication and sensation, such as vision or touch. 
A research problem or dataset is therefore characterized as \textit{multimodal} when it includes multiple such modalities.
In this paper we focus primarily, but not exclusively, on three modalities: natural \textit{language} which can be both written or spoken; \textit{visual} signals which are often represented with images or videos; and \textit{vocal} signals which encode sounds and para-verbal information such as prosody and vocal expressions. 



In order for Artificial Intelligence to make progress in understanding the world around us, it needs to be able to interpret and reason about multimodal messages. 
\textit{Multimodal machine learning} aims to build models that can process and relate information from multiple modalities. 
From early research on audio-visual speech recognition to the recent explosion of interest in language and vision models, multimodal machine learning is a vibrant multi-disciplinary field of increasing importance and with extraordinary potential.

 \begin{table*}[t]
\centering
\caption{A summary of applications enabled by multimodal machine learning. For each application area we identify the core technical challenges that need to be addressed in order to tackle it.}
 \label{tab:intro}
\begin{tabular}{|l|c|c|c|c|c|}
\hline
& \multicolumn{5}{c|}{\textsc{Challenges}} \\
  \hline
  \textsc{Applications} & \textsc{Representation}& \textsc{Translation} & \textsc{Alignment} & \textsc{Fusion} & \textsc{Co-learning}\\
  \hline
  \textbf{Speech recognition and synthesis} & & & & & \\
  \hspace{0.2cm}Audio-visual speech recognition & \checkmark & & \checkmark  & \checkmark  & \checkmark  \\  
  \hspace{0.2cm}(Visual) speech synthesis & \checkmark & \checkmark &  &  &  \\  
  \hline
  \textbf{Event detection} & & & & & \\
  \hspace{0.2cm}Action classification &\checkmark & & & \checkmark  & \checkmark \\
  \hspace{0.2cm}Multimedia event detection & \checkmark & & & \checkmark  & \checkmark \\
  \hline
  \textbf{Emotion and affect} & & & & & \\
  \hspace{0.2cm}Recognition & \checkmark & & \checkmark  & \checkmark &  \checkmark \\
  \hspace{0.2cm}Synthesis & \checkmark & \checkmark & & & \\
  \hline
  \textbf{Media description} & & & & & \\
  \hspace{0.2cm}Image description & \checkmark  & \checkmark & \checkmark & &  \checkmark \\
  \hspace{0.2cm}Video description & \checkmark & \checkmark & \checkmark & \checkmark &  \checkmark \\
  \hspace{0.2cm}Visual question-answering & \checkmark & & \checkmark & \checkmark &  \checkmark \\
  \hspace{0.2cm}Media summarization & \checkmark & \checkmark & &\checkmark & \\
  \hline
  \textbf{Multimedia retrieval}& & & & & \\
  \hspace{0.2cm}Cross modal retrieval & \checkmark  & \checkmark &  \checkmark & &  \checkmark \\
  \hspace{0.2cm}Cross modal hashing & \checkmark  &  & & &  \checkmark  \\
\hline
\end{tabular}
\end{table*}

The research field of Multimodal Machine Learning brings some unique challenges for computational researchers given the heterogeneity of the data. 
Learning from multimodal sources offers the possibility of capturing correspondences between modalities and gaining an in-depth understanding of natural phenomena. 
In this paper we identify and explore five core technical challenges (and related sub-challenges) surrounding multimodal machine learning. 
They are central to the multimodal setting and need to be tackled in order to progress the field. 
Our taxonomy goes beyond the typical early and late fusion split, and consists of the five following challenges:
\begin{itemize}[noitemsep,topsep=0pt,partopsep=0pt,leftmargin=12pt]
\item[1)] \textbf{Representation} A first fundamental challenge is learning how to represent and summarize multimodal data in a way that exploits the complementarity and redundancy of multiple modalities. 
The heterogeneity of multimodal data makes it challenging to construct such representations. 
For example, language is often symbolic while audio and visual modalities will be represented as signals. 
\item[2)] \textbf{Translation} A second challenge addresses how to translate (map) data from one modality to another. Not only is the data heterogeneous, but the relationship between modalities is often open-ended or subjective. For example, there exist a number of \emph{correct} ways to describe an image and and one perfect translation may not exist. 
\item[3)] \textbf{Alignment} A third challenge is to identify the direct relations between (sub)elements from two or more different modalities. For example, we may want to align the steps in a recipe to a video showing the dish being made. 
To tackle this challenge we need to measure similarity between different modalities and deal with possible long-range dependencies and ambiguities. 
\item[4)] \textbf{Fusion} A fourth challenge is to join information from two or more modalities to perform a prediction. For example, for audio-visual speech recognition, the visual description of the lip motion is fused with the speech signal to predict spoken words. The information coming from different modalities may have varying predictive power and noise topology, with possibly missing data in at least one of the modalities.
\item[5)] \textbf{Co-learning} A fifth challenge is to transfer knowledge between modalities, their representation, and their predictive models. This is exemplified by algorithms of co-training, conceptual grounding, and zero shot learning. Co-learning explores how knowledge learning from one modality can help a computational model trained on a different modality. This challenge is particularly relevant when one of the modalities has limited resources (e.g., annotated data). 
\end {itemize}
For each of these five challenges, we defines taxonomic classes and sub-classes to help structure the recent work in this emerging research field of multimodal machine learning. 
We start with a discussion of main applications of multimodal machine learning (Section \ref{sec:applications}) followed by a discussion on the recent developments on all of the five core technical challenges facing multimodal machine learning: representation (Section \ref{sec:rep}), translation (Section \ref{sec:translation}), alignment (Section \ref{sec:alignment}), fusion (Section \ref{sec:fusion}), and co-learning (Section \ref{sec:colearning}). 
We conclude with a discussion in Section \ref{sec:conclusion}.

\section{Applications: a historical perspective}
\label{sec:app}
\label{sec:applications}

Multimodal machine learning enables a wide range of applications: from audio-visual speech recognition to image captioning. 
In this section we present a brief history of multimodal applications, from its beginnings in audio-visual speech recognition to a recently renewed interest in language and vision applications. 

One of the earliest examples of multimodal research is audio-visual speech recognition (AVSR)  \cite{Yuhas1989}. 
It was motivated by the McGurk effect \cite{McGurk1976} --- an interaction between hearing and vision during speech perception. 
When human subjects heard the syllable /ba-ba/ while watching the lips of a person saying /ga-ga/, they perceived a third sound: /da-da/. 
These results motivated many researchers from the speech community to extend their approaches with visual information. 
Given the prominence of hidden Markov models (HMMs) in the speech community at the time \cite{Juang1991}, it is without surprise that many of the early models for AVSR were based on various HMM extensions \cite{Bourlard1996, Brand1997}.
While research into AVSR is not as common these days, it has seen renewed interest from the deep learning community \cite{Ngiam2011}.

While the original vision of AVSR was to improve speech recognition performance (e.g., word error rate) in all contexts, the experimental results showed that the main advantage of visual information was when the speech signal was noisy (i.e., low signal-to-noise ratio) \cite{Yuhas1989,Gurban2008,Ngiam2011}. 
In other words, the captured interactions between modalities were supplementary rather than complementary. 
The same information was captured in both, improving the robustness of the multimodal models but not improving the speech recognition performance in noiseless scenarios. 

A second important category of multimodal applications comes from the field of multimedia content indexing and retrieval \cite{Snoek2005, Atrey2010}. 
With the advance of personal computers and the internet, the quantity of digitized multimedia content has increased dramatically \cite{ytStats}.
While earlier approaches for indexing and searching these multimedia videos were keyword-based \cite{Snoek2005}, new research problems emerged when trying to search the visual and multimodal content directly. 
This led to new research topics in multimedia content analysis such as automatic shot-boundary detection \cite{Lienhart1998} and video summarization \cite{Evangelopoulos2013}. 
These research projects were supported by the TrecVid initiative from the National Institute of Standards and Technologies which introduced many high-quality datasets, including the multimedia event detection (MED) tasks started in 2011 \cite{trecvid}.


A third category of applications was established in the early 2000s around the emerging field of multimodal interaction with the goal of understanding human multimodal behaviors during social interactions. 
One of the first landmark datasets collected in this field is the AMI Meeting Corpus which contains more than 100 hours of video recordings of meetings, all fully transcribed and annotated \cite{Carletta2005}. 
Another important dataset is the SEMAINE corpus which allowed to study interpersonal dynamics between speakers and listeners \cite{McKeown2010}. 
This dataset formed the basis of the first audio-visual emotion challenge (AVEC) organized in 2011 \cite{Schuller2011}. 
The fields of emotion recognition and affective computing bloomed in the early 2010s thanks to strong technical advances in automatic face detection, facial landmark detection, and facial expression recognition \cite{DelaTorre2011}. 
The AVEC challenge continued annually afterward with the later instantiation including healthcare applications such as automatic assessment of depression and anxiety \cite{Valstar2013}. 
A great summary of recent progress in multimodal affect recognition was published by D'Mello et al. \cite{Dmello2015}. 
Their meta-analysis revealed that a majority of recent work on multimodal affect recognition show improvement when using more than one modality,  but this improvement is reduced when recognizing naturally-occurring emotions. 

Most recently, a new category of multimodal applications emerged with an emphasis on language and vision: media description. 
One of the most representative applications is image captioning where the task is to generate a text description of the input image \cite{Hodosh2013}. 
This is motivated by the ability of such systems to help the visually impaired in their daily tasks \cite{Bigham2010}. 
The main challenges media description is evaluation: how to evaluate the quality of the predicted descriptions. 
The task of visual question-answering (VQA) was recently proposed to address some of the evaluation challenges \cite{antol2015vqa}, where the goal is to answer a specific question about the image. 

In order to bring some of the mentioned applications to the real world we need to address a number of technical challenges facing multimodal machine learning. 
We summarize the relevant technical challenges for the above mentioned application areas in Table~\ref{tab:intro}.
One of the most important challenges is multimodal representation, the focus of our next section.

\section{Multimodal Representations}
\label{sec:rep}
\begin{figure*}[ht]
\centering
\subfloat[Joint representation\label{fig:joint-rep}]{\includegraphics[height=0.21\linewidth]{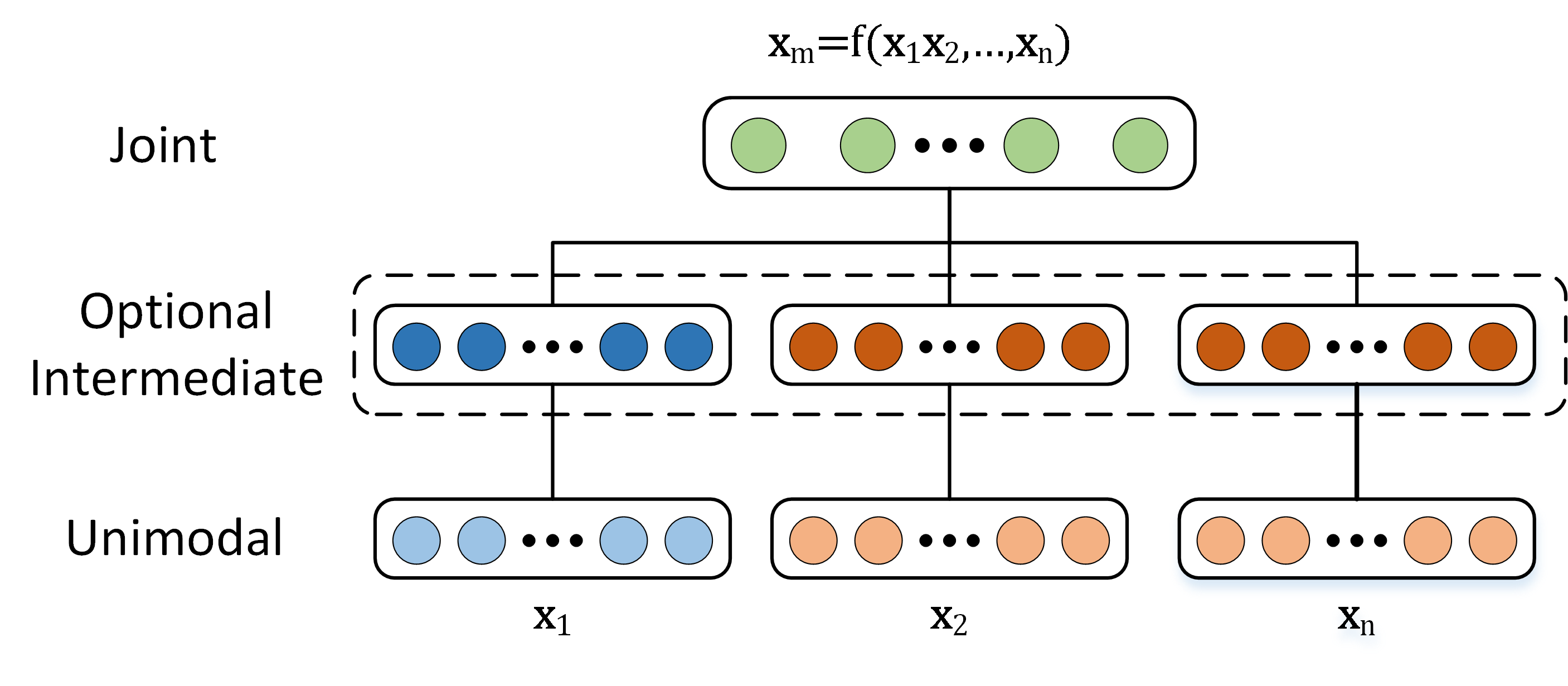}}
\hspace{0.5cm}
\subfloat[Coordinated representations \label{fig:coord-rep}]{\includegraphics[height=0.21\linewidth]{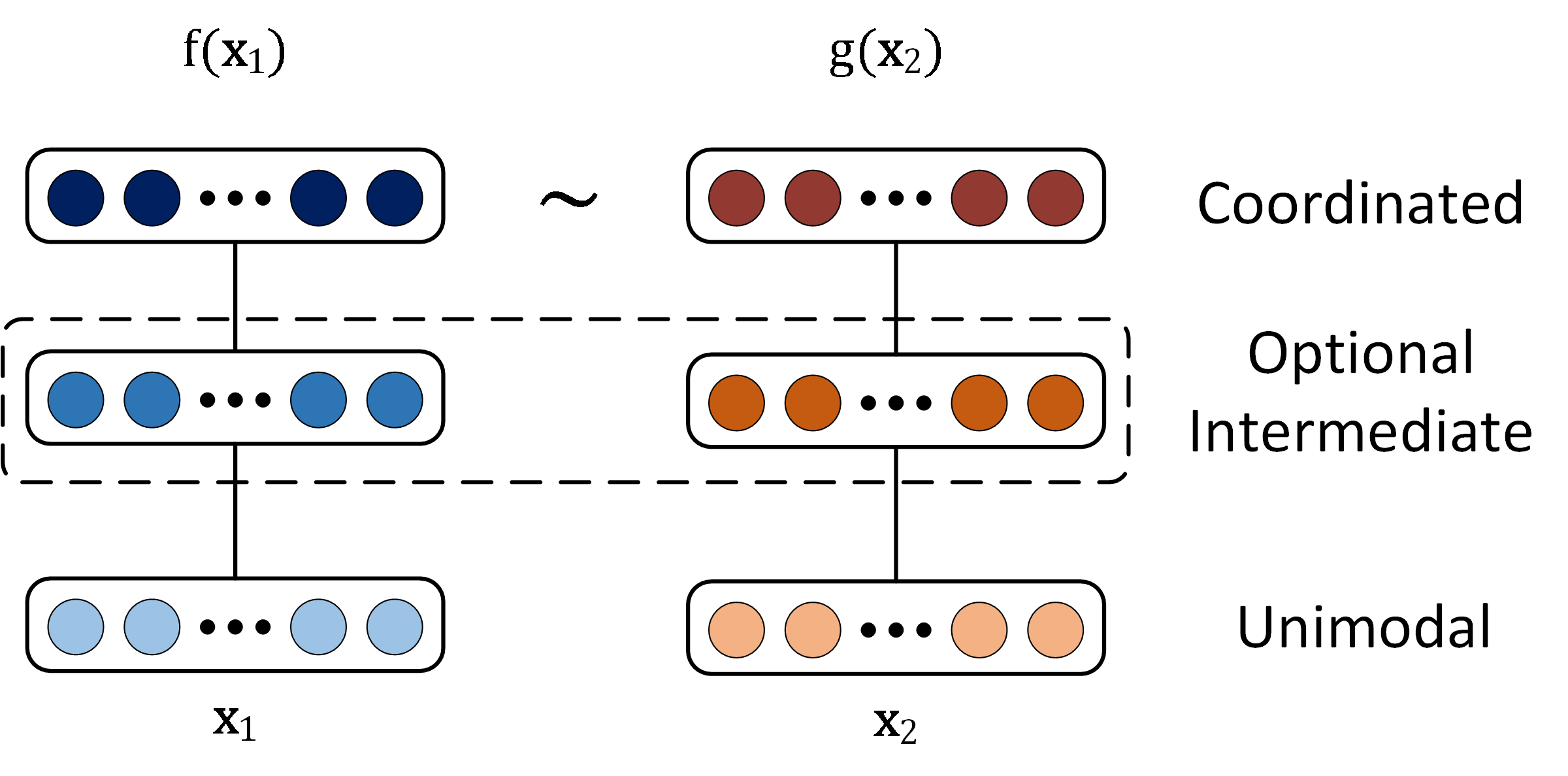}}
\caption{\label{fig:representation} Structure of \emph{joint} and \emph{coordinated} representations. Joint representations are projected to the same space using all of the modalities as input. Coordinated representations, on the other hand, exist in their own space, but are coordinated through a similarity (e.g. Euclidean distance) or structure constraint (e.g. partial order).}
\end{figure*}

Representing raw data in a format that a computational model can work with has always been a big challenge in machine learning. 
Following the work of Bengio et al. \cite{bengio2013representation} we use the term feature and representation interchangeably, with each referring to a vector or tensor representation of an entity, be it an image, audio sample, individual word, or a sentence. 
A multimodal representation is a representation of data using information from multiple such entities. 
Representing multiple modalities poses many difficulties: how to combine the data from heterogeneous sources; how to deal with different levels of noise; and how to deal with missing data.
The ability to represent data in a meaningful way is crucial to multimodal problems, and forms the backbone of any model. 

Good representations are important for the performance of machine learning models, as evidenced behind the recent leaps in performance of speech recognition \cite{Hinton2012} and visual object classification \cite{Krizhevsky2012} systems. 
Bengio et al. \cite{bengio2013representation} identify a number of properties for good representations: smoothness, temporal and spatial coherence, sparsity, and natural clustering amongst others. 
Srivastava and Salakhutdinov \cite{srivastava2012multimodal} identify additional desirable properties for multimodal representations: similarity in the representation space should reflect the similarity of the corresponding concepts, the representation should be easy to obtain even in the absence of some modalities, and finally, it should be possible to fill-in missing modalities given the observed ones.

The development of unimodal representations has been extensively studied \cite{Anagnostopoulos2012,bengio2013representation, Li2015}. 
In the past decade there has been a shift from hand-designed for specific applications to data-driven. 
For example, one of the most famous image descriptors in the early 2000s, the scale invariant feature transform (SIFT) was hand designed \cite{Lowe2004}, but currently most visual descriptions are learned from data using neural architectures such as convolutional neural networks (CNN) \cite{Krizhevsky2012}. 
Similarly, in the audio domain, acoustic features such as Mel-frequency cepstral coefficients (MFCC) have been superseded by data-driven deep neural networks in speech recognition \cite{Hinton2012} and recurrent neural networks for para-linguistic analysis \cite{Trigeorgis2016a}. 
In natural language processing, the textual features initially relied on counting word occurrences in documents, but have been replaced data-driven word embeddings that exploit the word context  \cite{mikolov2013distributed}. 
While there has been a huge amount of work on unimodal representation, up until recently most multimodal representations involved simple concatenation of unimodal ones \cite{Dmello2015}, but this has been rapidly changing.

To help understand the breadth of work, we propose two categories of multimodal representation: \textit{joint} and \textit{coordinated}. 
Joint representations combine the unimodal signals into the same representation space, while coordinated representations process unimodal signals separately, but enforce certain similarity constraints on them to bring them to what we term a coordinated space. An illustration of different multimodal representation types can be seen in Figure \ref{fig:representation}.

Mathematically, the joint representation is expressed as:
\begin{equation}
\label{eq:joint}
\mathbf{x}_m = f(\mathbf{x}_1,\ldots, \mathbf{x}_n),
\end{equation}
where the multimodal representation $\mathbf{x}_m$ is computed using function $f$ (e.g., a deep neural network, restricted Boltzmann machine, or a recurrent neural network) that relies on unimodal representations $\mathbf{x}_1,\ldots \mathbf{x}_n$. While coordinated representation is as follows:
\begin{equation}
f(\mathbf{x}_1)  \sim g(\mathbf{x}_2),
\end{equation}
where each modality has a corresponding projection function ($f$ and $g$ above) that maps it into a coordinated multimodal space. 
While the projection into the multimodal space is independent for each modality, but the resulting space is coordinated between them (indicated as $\sim$). 
Examples of such coordination include minimizing cosine distance \cite{Frome2013}, maximizing correlation \cite{andrew2013deep}, and enforcing a partial order \cite{Vendrov2016} between the resulting spaces.


\subsection{Joint Representations}
\label{sec:joint_rep}

We start our discussion with joint representations that project unimodal representations together into a multimodal space (Equation \ref{eq:joint}). 
Joint representations are mostly (but not exclusively) used in tasks where multimodal data is present both during training and inference steps. 
The simplest example of a joint representation is a concatenation of individual modality features (also referred to as early fusion \cite{Dmello2015}). 
In this section we discuss more advanced methods for creating joint representations starting with neural networks, followed by graphical models and recurrent neural networks (representative works can be seen in Table \ref{tab:representation}).

\noindent\textbf{Neural networks} have become a very popular method for unimodal data representation \cite{bengio2013representation}. 
They are used to represent visual, acoustic, and textual data, and are increasingly used in the multimodal domain \cite{Ngiam2011,Wang2015,Ouyang2014}.
In this section we describe how neural networks can be used to construct a joint multimodal representation, how to train them, and what advantages they offer.

In general, neural networks are made up of successive building blocks of inner products followed by non-linear activation functions. 
In order to use a neural network as a way to represent data, it is first trained to perform a specific task (e.g., recognizing objects in images). 
Due to the multilayer nature of deep neural networks each successive layer is hypothesized to represent the data in a more abstract way \cite{bengio2013representation}, hence it is common to use the final or penultimate neural layers as a form of data representation.
To construct a multimodal representation using neural networks each modality starts with several individual neural layers followed by a hidden layer that projects the modalities into a joint space \cite{Wu2014a, Mroueh2015, antol2015vqa,Ouyang2014}. 
The joint multimodal representation is then be passed through multiple hidden layers itself or used directly for prediction.
Such models can be trained end-to-end --- learning both to represent the data and to perform a particular task.
This results in a close relationship between multimodal representation learning and multimodal fusion when using neural networks.


As neural networks require a lot of labeled training data, it is common to pre-train such representations using an autoencoder on unsupervised data  \cite{Hinton1993}. 
The model proposed by Ngiam et al. \cite{Ngiam2011} extended the idea of using autoencoders to the multimodal domain. 
They used stacked denoising autoencoders to represent each modality individually and then fused them into a multimodal representation using another autoencoder layer. 
Similarly, Silberer and Lapata \cite{Silberer2014} proposed to use a multimodal autoencoder for the task of semantic concept grounding (see Section \ref{sec:non-parallel}). 
In addition to using a reconstruction loss to train the representation they introduce a term into the loss function that uses the representation to predict object labels. 
It is also common to fine-tune the resulting representation on a particular task at hand as the representation constructed using an autoencoder is generic and not necessarily optimal for a specific task \cite{Wang2015}.



The major advantage of neural network based joint representations comes from their often superior performance and the ability to pre-train the representations in an unsupervised manner. 
The performance gain is, however, dependent on the amount of data available for training. 
One of the disadvantages comes from the model not being able to handle missing data naturally --- although there are ways to alleviate this issue \cite{Ngiam2011, Wang2015}. 
Finally, deep networks are often difficult to train \cite{Glorot10}, but the field is making progress in better training techniques \cite{Srivastava2014}.

\noindent\textbf{Probabilistic graphical models} are another popular way to construct representations through the use of latent random variables \cite{bengio2013representation}. 
In this section we describe how probabilistic graphical models are used to represent unimodal and multimodal data.


The most popular approaches for graphical-model based representation are deep Boltzmann machines (DBM) \cite{salakhutdinov2009deep}, that stack restricted Boltzmann machines (RBM) \cite{Hinton2006} as building blocks.
Similar to neural networks, each successive layer of a DBM is expected to represent the data at a higher level of abstraction. 
The appeal of DBMs comes from the fact that they do not need supervised data for training \cite{salakhutdinov2009deep}. 
As they are graphical models the representation of data is probabilistic, however it is possible to convert them to a deterministic neural network --- but this loses the generative aspect of the model \cite{salakhutdinov2009deep}.

Work by Srivastava and Salakhutdinov \cite{Srivastava2012a} introduced multimodal deep belief networks as a multimodal representation. 
Kim et al. \cite{Kim2013} used a deep belief network for each modality and then combined them into joint representation for audiovisual emotion recognition. 
Huang and Kingsbury \cite{Huang2013} used a similar model for AVSR, and Wu et al. \cite{Wu2014} for audio and skeleton joint based gesture recognition.

Multimodal deep belief networks have been extended to multimodal DBMs by Srivastava and Salakhutdinov \cite{srivastava2012multimodal}. 
Multimodal DBMs are capable of learning joint representations from multiple modalities by merging two or more undirected graphs using a binary layer of hidden units on top of them. 
They allow for the low level representations of each modality to influence each other after the joint training due to the undirected nature of the model. 

Ouyang et al. \cite{Ouyang2014} explore the use of multimodal DBMs for the task of human pose estimation from multi-view data. 
They demonstrate that integrating the data at a later stage --- after unimodal data underwent nonlinear transformations --- was beneficial for the model. 
Similarly, Suk et al. \cite{Suk2014} use multimodal DBM representation to perform Alzheimer's disease classification from positron emission tomography and magnetic resonance imaging data.

\begin{table}[t]
  \centering
  \caption{\label{tab:representation} A summary of multimodal representation techniques. We identify three subtypes of joint representations (Section \ref{sec:joint_rep}) and two subtypes of coordinated ones  (Section \ref{sec:coord-rep}). For modalities + indicates the modalities combined.}
  \begin{tabular}{|l|c|c|}
  \hline
  \textsc{Representation} & \textsc{Modalities} & \textsc{Reference}\\
  \hline  
  \hline  
  \textbf{Joint} & &\\
  \hline
  \hspace{0.3cm}Neural networks  & Images + Audio& \cite{Ngiam2011, Mroueh2015, Wu2014a} \\
 & Images + Text  & \cite{Silberer2014}\\
  \hline
  \hspace{0.3cm}Graphical models & Images + Text & \cite{srivastava2012multimodal}\\
  & Images + Audio & \cite{Kim2013} \\
  \hline
  \hspace{0.3cm}Sequential & Audio + Video & \cite{Kahou2015,Nicolaou2011}  \\ 
                          & Images + Text& \cite{Rajagopalan2016} \\
  \hline
  \hline
  \textbf{Coordinated} & & \\
  \hline
 \hspace{0.3cm}Similarity & Images + Text & \cite{Frome2013, Kiros2014} \\
                         & Video + Text  & \cite{Xu2015,Pan2016}\\
  \hline
  \hspace{0.3cm}Structured & Images + Text & \cite{Cao2016,Vendrov2016,Zhang2014} \\
   & Audio + Articulatory& \cite{wang2015deep}\\
  \hline

\end{tabular}
\end{table}

One of the big advantages of using multimodal DBMs for learning multimodal representations is their generative nature, which allows for an easy way to deal with missing data --- even if a whole modality is missing, the model has a natural way to cope. 
It can also be used to generate samples of one modality in the presence of the other one, or both modalities from the representation. 
Similar to autoencoders the representation can be trained in an unsupervised manner enabling the use of unlabeled data. 
The major disadvantage of DBMs is the difficulty of training them --- high computational cost, and the need to use approximate variational training methods \cite{srivastava2012multimodal}. 




\noindent\textbf{Sequential Representation}. So far we have discussed models that can represent fixed length data, however, we often need to represent varying length sequences such as sentences, videos, or audio streams. 
In this section we describe models that can be used to represent such sequences.

Recurrent neural networks (RNNs), and their variants such as  long-short term memory (LSTMs) networks \cite{hochreiter1997long}, have recently gained popularity due to their success in sequence modeling across various tasks \cite{Bahdanau2014,Venugopalan2015}. 
So far RNNs have mostly been used to represent unimodal sequences of words, audio, or images, with most success in the language domain. 
Similar to traditional neural networks, the hidden state of an RNN can be seen as a representation of the data, i.e., the hidden state of RNN at timestep $t$ can be seen as the summarization of the sequence up to that timestep. 
This is especially apparent in RNN encoder-decoder frameworks where the task of an encoder is to represent a sequence in the hidden state of an RNN in such a way that a decoder could reconstruct it \cite{Bahdanau2014}.

The use of RNN representations has not been limited to the unimodal domain. 
An early use of constructing a multimodal representation using RNNs comes from work by Cosi et al. \cite{Cosi1994} on AVSR. 
They have also been used for representing audio-visual data for affect recognition \cite{Nicolaou2011, Chen2015a} and to represent multi-view data such as different visual cues for human behavior analysis \cite{Rajagopalan2016}. 




\subsection{Coordinated Representations}
\label{sec:coord-rep}

An alternative to a joint multimodal representation is a coordinated representation. 
Instead of projecting the modalities together into a joint space, we learn separate representations for each modality but coordinate them through a constraint.
We start our discussion with coordinated representations that enforce similarity between representations, moving on to coordinated representations that enforce more structure on the resulting space (representative works of different coordinated representations can be seen in Table \ref{tab:representation}).

\noindent\textbf{Similarity models} minimize the distance between modalities in the coordinated space. 
For example such models encourage the representation of the word \emph{dog} and an image of a dog to have a smaller distance between them than distance between the word \emph{dog} and an image of a car \cite{Frome2013}.
One of the earliest examples of such a representation comes from the work by Weston et al. \cite{Weston2010,weston2011wsabie} on the WSABIE (web scale annotation by image embedding) model, where a coordinated space was constructed for images and their annotations. 
WSABIE constructs a simple linear map from image and textual features such that corresponding annotation and image representation would have a higher inner product (smaller cosine distance) between them than non-corresponding ones.

More recently, neural networks have become a popular way to construct coordinated representations, due to their ability to learn representations. 
Their advantage lies in the fact that they can jointly learn coordinated representations in an end-to-end manner.  
An example of such coordinated representation is DeViSE --- a deep visual-semantic embedding \cite{Frome2013}. 
DeViSE uses a similar inner product and ranking loss function to WSABIE but uses more complex image and word embeddings. 
Kiros et al. \cite{Kiros2014} extended this to sentence and image coordinated representation by using an LSTM model and a pairwise ranking loss to coordinate the feature space. 
Socher et al. \cite{Socher2014} tackle the same task, but extend the language model to a dependency tree RNN to incorporate compositional semantics. 
A similar model was also proposed by Pan et al.\cite{Pan2016}, but using videos instead of images. Xu et al. \cite{Xu2015} also constructed a coordinated space between videos and sentences using a $\langle$subject, verb, object$\rangle$ compositional language model and a deep video model. 
This representation was then used for the task of cross-modal retrieval and video description. 





While the above models enforced similarity between representations, \textbf{structured coordinated space} models go beyond that and enforce additional constraints between the modality representations. 
The type of structure enforced is often based on the application, with different constraints for hashing, cross-modal retrieval, and image captioning. 

Structured coordinated spaces are commonly used in cross-modal hashing --- compression of high dimensional data into compact binary codes with similar binary codes for similar objects  \cite{Wang2014a}. 
The idea of cross-modal hashing is to create such codes for cross-modal retrieval \cite{Kumar2011, Bronstein2010, Jiang2015}. 
Hashing enforces certain constraints on the resulting multimodal space: 1) it has to be an $N$-dimensional Hamming space --- a binary representation with controllable number of bits; 2) the same object from different modalities has to have a similar hash code; 3) the space has to be similarity-preserving. 
Learning how to represent the data as a hash function attempts to enforce all of these three requirements \cite{Kumar2011,Bronstein2010}.  
For example, Jiang and Li \cite{Jiang2016} introduced a method to learn such common binary space between sentence descriptions and corresponding images using end-to-end trainable deep learning techniques. 
While Cao et al. \cite{Cao2016} extended the approach with a more complex LSTM sentence representation and introduced an outlier insensitive bit-wise margin loss and a relevance feedback based semantic similarity constraint.
Similarly, Wang et al. \cite{Wang2016} constructed a coordinated space in which images (and sentences) with similar meanings are closer to each other. 

Another example of a structured coordinated representation comes from order-embeddings of images and language \cite{Vendrov2016,Zhang2016}. 
The model proposed by Vendrov et al. \cite{Vendrov2016} enforces a dissimilarity metric that is asymmetric and implements the notion of partial order in the multimodal space. 
The idea is to capture a partial order of the language and image representations --- enforcing a hierarchy on the space; for example image of “a woman walking her dog“ $\rightarrow$ text “woman walking her dog” $\rightarrow$ text “woman walking”. 
A similar model using denotation graphs was also proposed by Young et al. \cite{Young2014} where denotation graphs are used to induce a partial ordering.
Lastly, Zhang et al. present how exploiting structured representations of text and images can create concept taxonomies in an unsupervised manner \cite{Zhang2016}.

A special case of a structured coordinated space is one based on canonical correlation analysis (CCA) \cite{Hotelling1936}. 
CCA computes a linear projection which maximizes the correlation between two random variables (in our case modalities) and enforces orthogonality of the new space. 
CCA models have been used extensively for cross-modal retrieval \cite{Hardoon2003, Rasiwasia2010,Klein2015} and audiovisual signal analysis \cite{Sargin2007,Slaney2000}.
Extensions to CCA attempt to construct a correlation maximizing nonlinear projection \cite{lai2000kernel,andrew2013deep}. 
Kernel canonical correlation analysis (KCCA) \cite{lai2000kernel} uses reproducing kernel Hilbert spaces for projection. 
However, as the approach is nonparametric it scales poorly with the size of the training set and has issues with very large real-world datasets. 
Deep canonical correlation analysis (DCCA) \cite{andrew2013deep} was introduced as an alternative to KCCA and addresses the scalability issue, it was also shown to lead to better correlated representation space. 
Similar correspondence autoencoder \cite{Feng2014} and deep correspondence RBMs \cite{Feng2015} have also been proposed for cross-modal retrieval.

CCA, KCCA, and DCCA are unsupervised techniques and only optimize the correlation over the representations, thus mostly capturing what is shared across the modalities. 
Deep canonically correlated autoencoders \cite{wang2015deep} also include an autoencoder based data reconstruction term. 
This encourages the representation to also capture modality specific information. 
Semantic correlation maximization method \cite{Zhang2014} also encourages semantic relevance, while retaining correlation maximization and orthogonality of the resulting space --- this leads to a combination of CCA and cross-modal hashing techniques.

\subsection{Discussion}

In this section we identified two major types of multimodal representations --- joint and coordinated. 
Joint representations project multimodal data into a common space and are best suited for situations when all of the modalities are present during inference. 
They have been extensively used for AVSR, affect, and multimodal gesture recognition. 
Coordinated representations, on the other hand, project each modality into a separate but coordinated space, making them suitable for applications where only one modality is present at test time, such as: multimodal retrieval and translation (Section \ref{sec:translation}), grounding (Section \ref{sec:non-parallel}), and zero shot learning (Section \ref{sec:non-parallel}). 
Finally, while joint representations have been used in situations to construct representations of more than two modalities, coordinated spaces have, so far, been mostly limited to two modalities.

\section{Translation}
\label{sec:translation}
\label{sec:translation}

\begin{table}
\begin{center}
\caption{Taxonomy of multimodal translation research. For each class and sub-class, we include example tasks with references. Our taxonomy also includes the directionality of the translation:  unidirectional ($\Rightarrow$) and bidirectional ($\Leftrightarrow$). }
 \label{tab:translation}
 \begin{tabular}{|l|c|c|c|c}
\hline
  & \textsc{Tasks} & \textsc{Dir.}  & \textsc{References}\\
\hline
\hline
\textbf{Example-based} &  & & \\
\hline
 \hspace{0.1cm}Retrieval & Image captioning & $\Rightarrow$ & \cite{Ordonez2011,Farhadi2010}\\
  & Media retrieval& $\Leftrightarrow$   & \cite{Socher2014,Xu2015} \\
 & Visual speech  &  $\Rightarrow$  & \cite{Bregler1997}\\
 & Image captioning  &  $\Leftrightarrow$ & \cite{karpathy2014deep, Karpathy2015}\\
\hline
    \hspace{0.1cm}Combination & Image captioning& $\Rightarrow$  & \cite{Lebret2015,Kuznetsova2012,Gupta2012} \\
\hline
\hline
   \textbf{Generative} & & &\\
\hline 
\hspace{0.1cm}Grammar based  & Video description & $\Rightarrow$     & \cite{Barbu2012,Thomason2014}\\
  & Image description & $\Rightarrow$  & \cite{Li2011,Mitchell2012,Elliott2013}  \\
\hline
\hspace{0.1cm}Encoder-decoder   & Image captioning &  $\Rightarrow$  & \cite{Mao2015,Kiros2014} \\
      & Video description & $\Rightarrow$  & \cite{Venugopalan2015,yu2015video} \\
                     & Text to image  & $\Rightarrow$   & \cite{Mansimov2016, Reed2016}  \\
\hline
\hspace{0.1cm}Continuous & Sounds synthesis & $\Rightarrow$  & \cite{Owens2016,VandenOord2016} \\
                   & Visual speech & $\Rightarrow$   & \cite{Anderson2013,Deena2009,Taylor2012} \\
\hline
\end{tabular}
\end{center}
\end{table}

A big part of multimodal machine learning is concerned with translating (mapping) from one modality to another. 
Given an entity in one modality the task is to generate the same entity in a different modality. 
For example given an image we might want to generate a sentence describing it or given a textual description generate an image matching it.
Multimodal translation is a long studied problem, with early work in speech synthesis \cite{Hunt1996}, visual speech generation \cite{Masuko1998} video description \cite{Kojima2002}, and cross-modal retrieval \cite{Rasiwasia2010}.
 

More recently, multimodal translation has seen renewed interest due to combined efforts of the computer vision and natural language processing (NLP) communities \cite{Bernardi2016} and recent availability of large multimodal datasets \cite{Chen2015, Torabi2015}. 
A particularly popular problem is visual scene description, also known as image \cite{Vinyals2014} and video captioning \cite{Venugopalan2015}, which acts as a great test bed for a number of computer vision and NLP problems. 
To solve it, we not only need to fully understand the visual scene and to identify its salient parts, but also to produce grammatically correct and comprehensive yet concise sentences describing it.

\begin{figure*}[t]
\centering
\subfloat[Example-based \label{fig:example-based}]{\includegraphics[height=0.23\linewidth]{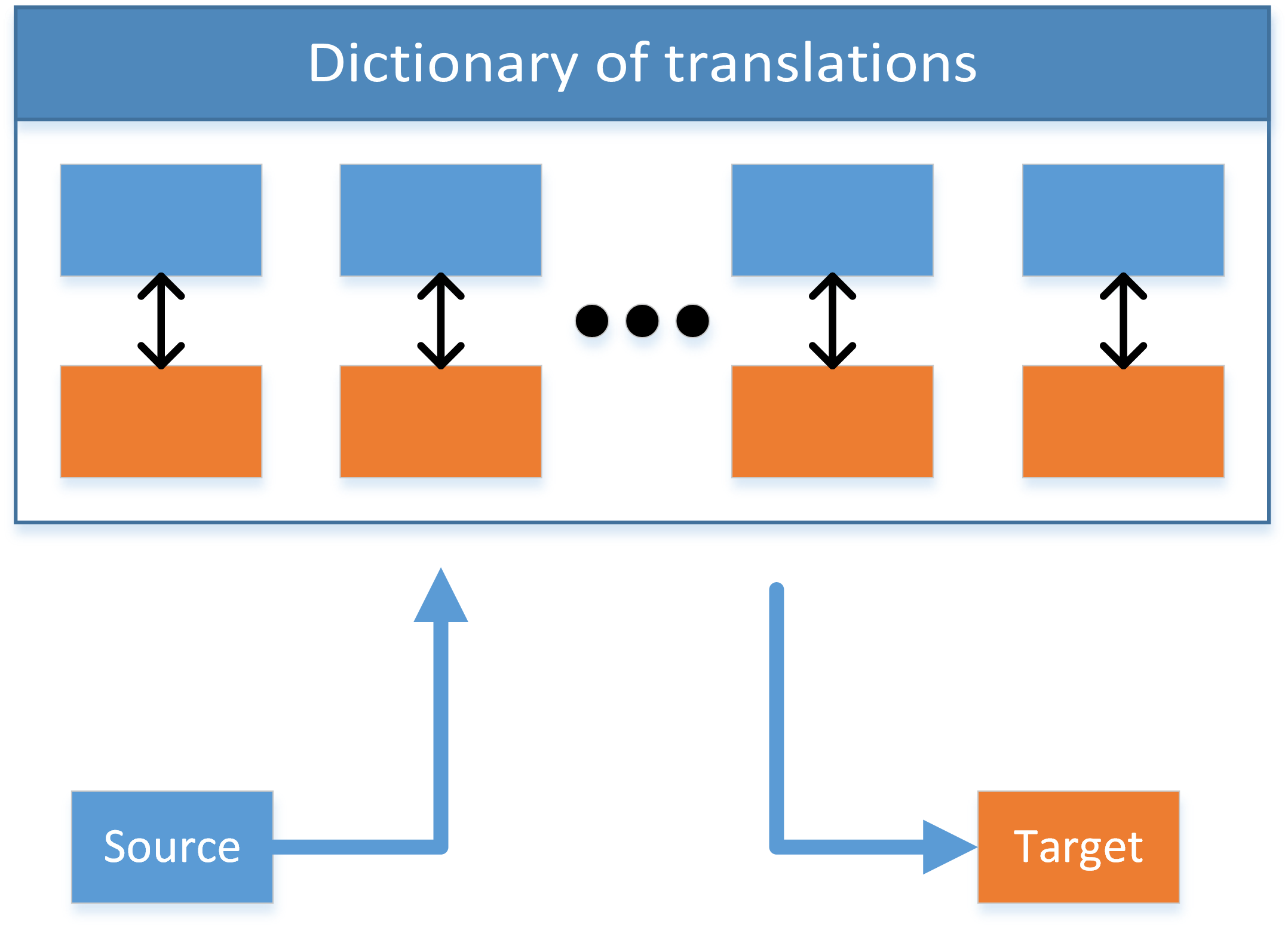}}
\hspace{2cm}
\subfloat[Generative \label{fig:generative}]{\includegraphics[height=0.23\linewidth]{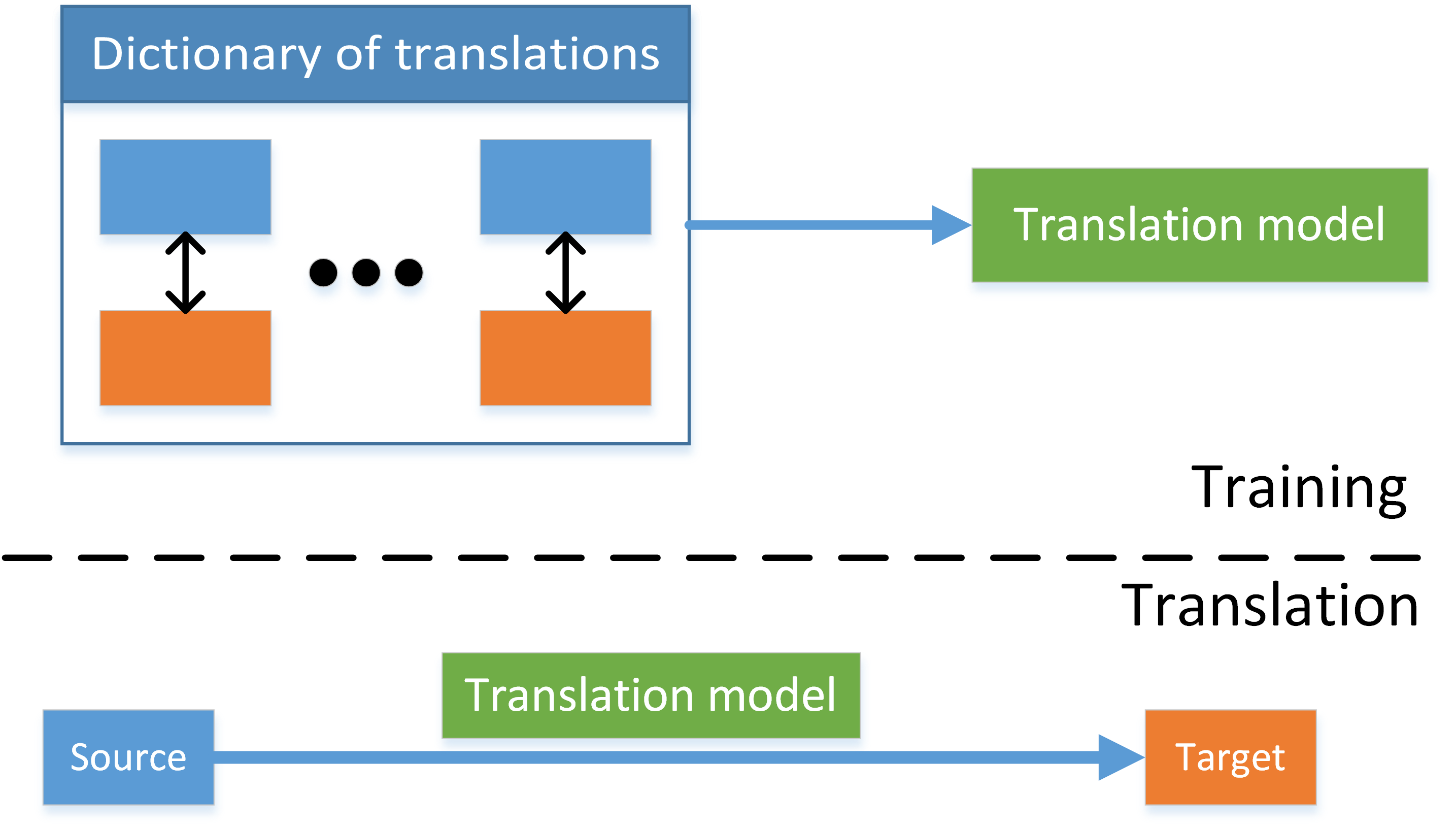}}
\caption{\label{fig:translation} Overview of \emph{example-based} and \emph{generative} multimodal translation. The former retrieves the best translation from a dictionary, while the latter first trains a translation model on the dictionary and then uses that model for translation.}
\end{figure*}

While the approaches to multimodal translation are very broad and are often modality specific, they share a number of unifying factors.
We categorize them into two types --- \textit{example-based}, and \textit{generative}. 
Example-based models use a \emph{dictionary} when translating between the modalities. 
Generative models, on the other hand, construct a \emph{model} that is able to produce a translation. 
This distinction is similar to the one between non-parametric and parametric machine learning approaches and is illustrated in Figure \ref{fig:translation}, with representative examples summarized in Table \ref{tab:translation}.


\emph{Generative} models are arguably more challenging to build as they require the ability to generate signals or sequences of symbols (e.g., sentences). 
This is difficult for any modality --- visual, acoustic, or verbal, especially when temporally and structurally consistent sequences need to be generated. 
This led to many of the early multimodal translation systems relying on \emph{example-based} translation. 
However, this has been changing with the advent of deep learning models that are capable of generating images \cite{Oord2016,Reed2016}, sounds \cite{Owens2016,VandenOord2016}, and text \cite{Bahdanau2014}.

\subsection{Example-based}

Example-based algorithms are restricted by their training data --- dictionary (see Figure \ref{fig:example-based}).
We identify two types of such algorithms: retrieval based, and combination based. 
\emph{Retrieval}-based models directly use the retrieved translation without modifying it, while \emph{combination}-based models rely on more complex rules to create translations based on a number of retrieved instances.

\noindent\textbf{Retrieval-based models} are arguably the simplest form of multimodal translation. 
They rely on finding the closest sample in the dictionary and using that as the translated result. 
The retrieval can be done in \emph{unimodal} space or intermediate \emph{semantic} space.

Given a source modality instance to be translated, unimodal retrieval finds the closest instances in the dictionary in the space of the source --- for example, visual feature space for images. 
Such approaches have been used for visual speech synthesis, by retrieving the closest matching visual example of the desired phoneme \cite{Bregler1997}. 
They have also been used in concatenative text-to-speech systems \cite{Hunt1996}.
More recently, Ordonez et al. \cite{Ordonez2011} used  unimodal retrieval to generate image descriptions by using global image features to retrieve caption candidates \cite{Ordonez2011}. 
Yagcioglu et al. \cite{Yagcioglu2015} used a CNN-based image representation to retrieve visually similar images using adaptive neighborhood selection.
Devlin et al. \cite{Devlin2015} demonstrated that a simple $k$-nearest neighbor retrieval with consensus caption selection achieves competitive translation results when compared to more complex generative approaches.
The advantage of such unimodal retrieval approaches is that they only require the representation of a single modality through which we are performing retrieval.
However, they often require an extra processing step such as re-ranking of retrieved translations \cite{Ordonez2011,Mason2014, Yagcioglu2015}. 
This indicates a major problem with this approach --- similarity in unimodal space does not always imply a good translation.

An alternative is to use an intermediate semantic space for similarity comparison during retrieval. 
An early example of a hand crafted semantic space is one used by Farhadi et al. \cite{Farhadi2010}. 
They map both sentences and images to a space of $\langle$object, action, scene$\rangle$, retrieval of relevant caption to an image is then performed in that space.
In contrast to hand-crafting a representation, Socher et al. \cite{Socher2014} learn a coordinated representation of sentences and CNN visual features (see Section \ref{sec:coord-rep} for description of coordinated spaces). 
They use the model for both translating from text to images and from images to text. 
Similarly, Xu et al. \cite{Xu2015} used a coordinated space of videos and their descriptions for cross-modal retrieval. 
Jiang and Li \cite{Jiang2015} and Cao et al. \cite{Cao2016} use cross-modal hashing to perform multimodal translation from images to sentences and back, while Hodosh et al. \cite{Hodosh2013} use a multimodal KCCA space for image-sentence retrieval. 
Instead of aligning images and sentences globally in a common space, Karpathy et al. \cite{karpathy2014deep} propose a multimodal similarity metric that internally aligns image fragments (visual objects) together with sentence fragments (dependency tree relations).

Retrieval approaches in semantic space tend to perform better than their unimodal counterparts as they are retrieving examples in a more meaningful space that reflects both modalities and that is often optimized for retrieval. 
Furthermore, they allow for bi-directional translation, which is not straightforward with unimodal methods.
However, they require manual construction or learning of such a semantic space, which often relies on the existence of large training dictionaries (datasets of paired samples).

\noindent\textbf{Combination-based models} take the retrieval based approaches one step further. 
Instead of just retrieving examples from the dictionary, they combine them in a meaningful way to construct a better translation. 
Combination based media description approaches are motivated by the fact that sentence descriptions of images share a common and simple structure that could be exploited.
Most often the rules for combinations are hand crafted or based on heuristics. 

Kuznetsova et al. \cite{Kuznetsova2012} first retrieve phrases that describe visually similar images and then combine them to generate novel descriptions of the query image by using Integer Linear Programming with a number of hand crafted rules. 
Gupta et al. \cite{Gupta2012} first find $k$ images most similar to the source image, and then use the phrases extracted from their captions to generate a target sentence. 
Lebret et al. \cite{Lebret2015} use a CNN-based image representation to infer phrases that describe it.
The predicted phrases are then combined using a trigram constrained language model.

A big problem facing example-based approaches for translation is that the model is the entire dictionary --- making the model large and inference slow (although, optimizations such as hashing alleviate this problem). 
Another issue facing example-based translation is that it is unrealistic to expect that a single comprehensive and accurate translation relevant to the source example will always exist in the dictionary --- unless the task is simple or the dictionary is very large.
This is partly addressed by combination models that are able to construct more complex structures. 
However, they are only able to perform translation in one direction, while semantic space retrieval-based models are able to perform it both ways. 



\subsection{Generative approaches}
Generative approaches to multimodal translation construct models that can perform multimodal translation given a unimodal source instance. 
It is a challenging problem as it requires the ability to both understand the source modality and to generate the target sequence or signal.
As discussed in the following section, this also makes such methods much more difficult to evaluate, due to large space of possible correct answers.

In this survey we focus on the generation of three modalities: language, vision, and sound. 
Language generation has been explored for a long time \cite{Ratnaparkhi2000}, with a lot of recent attention for tasks such as image and video description \cite{Bernardi2016}. 
Speech and sound generation has also seen a lot of work with a number of historical \cite{Hunt1996} and modern approaches \cite{Owens2016,VandenOord2016}.
Photo-realistic image generation has been less explored, and is still in early stages \cite{Mansimov2016,Reed2016}, however, there have been a number of attempts at generating abstract scenes \cite{Zitnick2013}, computer graphics \cite{Coyne2001}, and talking heads \cite{Anderson2013}.

We identify three broad categories of generative models: \emph{grammar-based}, \emph{encoder-decoder}, and \emph{continuous generation} models. Grammar based models simplify the task by restricting the target domain by using a grammar, e.g., by generating restricted sentences based on a $\langle$subject, object, verb$\rangle$ template. Encoder-decoder models first encode the source modality to a latent representation which is then used by a decoder to generate the target modality. Continuous generation models generate the target modality continuously based on a stream of source modality inputs and are most suited for translating between temporal sequences --- such as text-to-speech.

\noindent\textbf{Grammar-based models} rely on a pre-defined grammar for generating a particular modality. 
They start by detecting high level concepts from the source modality, such as objects in images and actions from videos. 
These detections are then incorporated together with a generation procedure based on a pre-defined grammar to result in a target modality.

Kojima et al. \cite{Kojima2002} proposed a system to describe human behavior in a video using  the detected position of the person's head and hands and rule based natural language generation that incorporates a hierarchy of concepts and actions.
Barbu et al. \cite{Barbu2012} proposed a video description model that generates sentences of the form: \emph{who} did \emph{what} to \emph{whom} and \emph{where} and \emph{how} they did it. 
The system was based on handcrafted object and event classifiers and used a restricted grammar suitable for the task.
Guadarrama et al. \cite{Guadarrama2013} predict $\langle$subject, verb, object$\rangle$ triplets describing a video using semantic hierarchies that use more general words in case of uncertainty. 
Together with a language model their approach allows for translation of verbs and nouns not seen in the dictionary.

To describe images, Yao et al. \cite{Yao2010} propose to use an and-or graph-based model together with domain-specific lexicalized grammar rules, targeted visual representation scheme, and a hierarchical knowledge ontology. 
Li et al. \cite{Li2011} first detect objects, visual attributes, and spatial relationships between objects. 
They then use an $n$-gram language model on the visually extracted phrases to generate $\langle$subject, preposition, object$\rangle$ style sentences.
Mitchell et al. \cite{Mitchell2012} use a more sophisticated tree-based language model to generate syntactic trees instead of filling in templates, leading to more diverse descriptions. 
A majority of approaches represent the whole image jointly as a bag of visual objects without capturing their spatial and semantic relationships. 
To address this, Elliott et al. \cite{Elliott2013} propose to explicitly model proximity relationships of objects for image description generation.


Some grammar-based approaches rely on graphical models to generate the target modality. 
An example includes BabyTalk \cite{Kulkarni2013}, which given an image generates $\langle$object, preposition, object$\rangle$ triplets, that are used together with a conditional random field to construct the sentences. 
Yang et al. \cite{Yang2011} predict a set of $\langle$noun, verb, scene, preposition$\rangle$ candidates using visual features extracted from an image and combine them into a sentence using a statistical language model and hidden Markov model style inference. 
A similar approach has been proposed by Thomason et al. \cite{Thomason2014}, where a factor graph model is used for video description of the form $\langle$subject, verb, object, place$\rangle$. 
The factor model exploits language statistics to deal with noisy visual representations. 
Going the other way Zitnick et al. \cite{Zitnick2013} propose to use conditional random fields to generate abstract visual scenes based on language triplets extracted from sentences.

An advantage of grammar-based methods is that they are more likely to generate syntactically (in case of language) or logically correct target instances as they use predefined templates and restricted grammars. 
However, this limits them to producing formulaic rather than creative translations.
Furthermore, grammar-based methods rely on complex pipelines for concept detection, with each concept requiring a separate model and a separate training dataset. 

\noindent\textbf{Encoder-decoder models} based on end-to-end  trained neural networks are currently some of the most popular techniques for multimodal translation.  
The main idea behind the model is to first encode a source modality into a vectorial representation and then to use a decoder module to generate the target modality, all this in a single pass pipeline. 
Although, first used for machine translation \cite{Kalchbrenner2013}, such models have been successfully used for image captioning \cite{Mao2015, Vinyals2014}, and video description \cite{Venugopalan2015,rohrbach2015long}. 
So far, encoder-decoder models have been mostly used to generate text, but they can also be used to generate images \cite{Mansimov2016,Reed2016}, and continuos generation of speech and sound \cite{Owens2016,VandenOord2016}.

The first step of the encoder-decoder model is to encode the source object, this is done in modality specific way.
Popular models to encode acoustic signals include RNNs \cite{Chan2016} and DBNs \cite{Hinton2012}. 
Most of the work on encoding words sentences uses distributional semantics \cite{mikolov2013distributed} and variants of RNNs \cite{Bahdanau2014}.
Images are most often encoded using convolutional neural networks (CNN) \cite{Krizhevsky2012, Simonyan2015}. 
While learned CNN representations are common for encoding images, this is not the case for videos where hand-crafted features are still commonly used \cite{rohrbach2015long, Thomason2014}.
While it is possible to use unimodal representations to encode the source modality, it has been shown that using a coordinated space (see Section \ref{sec:coord-rep}) leads to better results  \cite{Kiros2014, Pan2016, Xu2015}.

Decoding is most often performed by an RNN or an LSTM using the encoded representation as the initial hidden state \cite{Vinyals2014, vinyals2015show, Fan2014, Mansimov2016}. 
A number of extensions have been proposed to traditional LSTM models to aid in the task of translation. 
A guide vector could be used to tightly couple the solutions in the image input \cite{Jia2015}.
Venugopalan et al. \cite{Venugopalan2015} demonstrate that it is beneficial to pre-train a decoder LSTM for image captioning before fine-tuning it to video description. 
Rohrbach et al. \cite{rohrbach2015long} explore the use of various LSTM architectures (single layer, multilayer, factored) and a number of training and regularization techniques for the task of video description.

A problem facing translation generation using an RNN is that the model has to generate a description from a single vectorial representation of the image, sentence, or video. 
This becomes especially difficult when generating long sequences as these models tend to \emph{forget} the initial input. 
This has been partly addressed by neural attention models (see Section \ref{sec:implicit-alignment}) that allow the network to focus on certain parts of an image \cite{xu2015show}, sentence \cite{Bahdanau2014}, or video  \cite{Yao2015} during generation.

Generative attention-based RNNs have also been used for the task of generating images from sentences \cite{Mansimov2016}, while the results are still far from photo-realistic they show a lot of promise. 
More recently, a large amount of progress has been made in generating images using generative adversarial networks \cite{Goodfellow2014}, which have been used as an alternative to RNNs for image generation from text \cite{Reed2016}.


While neural network based encoder-decoder systems have been very successful they still face a number of issues. 
Devlin et al. \cite{Devlin2015} suggest that it is possible that the network is \emph{memorizing} the training data rather than learning how to understand the visual scene and generate it. 
This is based on the observation that $k$-nearest neighbor models perform very similarly to those based on generation.
Furthermore, such models often require large quantities of data for training.

\noindent\textbf{Continuous generation models} are intended for sequence translation and produce outputs at every timestep in an online manner.
These models are useful when translating from a sequence to a sequence such as text to speech, speech to text, and video to text. 
A number of different techniques have been proposed for such modeling --- graphical models, continuous encoder-decoder approaches, and various other regression or classification techniques. 
The extra difficulty that needs to be tackled by these models is the requirement of temporal consistency between modalities.

A lot of early work on sequence to sequence translation used graphical or latent variable models. 
Deena and Galata \cite{Deena2009} proposed to use a shared Gaussian process latent variable model for audio-based visual speech synthesis. 
The model creates a shared latent space between audio and visual features that can be used to generate one space from the other, while enforcing  temporal consistency of visual speech at different timesteps. 
Hidden Markov models (HMM) have also been used for visual speech generation \cite{Taylor2012} and text-to-speech \cite{Zen2009} tasks.
They have also been extended to use cluster adaptive training to allow for training on multiple speakers, languages, and emotions allowing for more control when generating speech signal \cite{Zen2012} or visual speech parameters \cite{Anderson2013}.

Encoder-decoder models have recently become popular for sequence to sequence modeling. 
Owens et al. \cite{Owens2016} used an LSTM to generate sounds resulting from drumsticks based on video. 
While their model is capable of generating sounds by predicting a cochleogram from CNN visual features, they found that retrieving a closest audio sample based on the predicted cochleogram led to best results.
Directly modeling the raw audio signal for speech and music generation has been proposed by van den Oord et al. \cite{VandenOord2016}. 
The authors propose using hierarchical fully convolutional neural networks, which show a large improvement over previous state-of-the-art for the task of speech synthesis. 
RNNs have also been used for speech to text translation (speech recognition) \cite{graves2013speech}.
More recently encoder-decoder based continuous approach was shown to be good at predicting letters from a speech signal represented as a filter bank spectra \cite{Chan2016} --- allowing for more accurate recognition of rare and out of vocabulary words.
Collobert et al. \cite{Collobert2016} demonstrate how to use a raw audio signal directly for speech recognition, eliminating the need for audio features.

A lot of earlier work used graphical models for multimodal translation between continuous signals. 
However, these methods are being replaced by neural network encoder-decoder based techniques. 
Especially as they have recently been shown to be able to represent and generate complex visual and acoustic signals.

\subsection{Model evaluation and discussion}
\label{sec:translation-evaluation}

A major challenge facing multimodal translation methods is that they are very difficult to evaluate. 
While some tasks such as speech recognition have a single correct translation, tasks such as speech synthesis and media description do not.
Sometimes, as in language translation, multiple answers are correct and deciding which translation is better is often subjective. 
Fortunately, there are a number of approximate automatic metrics that aid in model evaluation.

Often the ideal way to evaluate a subjective task is through human judgment. 
That is by having a group of people evaluating each translation. 
This can be done on a Likert scale where each translation is evaluated on a certain dimension: naturalness and mean opinion score for speech synthesis \cite{Zen2012, VandenOord2016}, realism for visual speech synthesis \cite{Anderson2013, Taylor2012}, and grammatical and semantic correctness, relevance, order, and detail for media description \cite{Kulkarni2013,Mitchell2012, Chen2015, Venugopalan2015}. 
Another option is to perform preference studies where two (or more) translations are presented to the participant for preference comparison \cite{Zen2012, Taylor2012}. 
However, while user studies will result in evaluation closest to human judgments they are time consuming and costly. 
Furthermore, they require care when constructing and conducting them to avoid fluency, age, gender and culture biases.

While human studies are a gold standard for evaluation, a number of automatic alternatives have been proposed for the task of media description: BLEU \cite{Papineni2002}, ROUGE \cite{Lin2003}, Meteor \cite{Denkowski2014}, and CIDEr \cite{Vedantam2015}. 
These metrics are directly taken from (or are based on) work in machine translation and compute a score that measures the similarity between the generated and ground truth text. 
However, the use of them has faced a lot of criticism. 
Elliott and Keller \cite{Elliott2014} showed that sentence-level unigram BLEU is only weakly correlated with human judgments.
Huang et al. \cite{huang2016visual} demonstrated that the correlation between human judgments and BLEU and Meteor is very low for visual story telling task.
Furthermore, the ordering of approaches based on human judgments did not match that of the ordering using automatic metrics on the MS COCO challenge \cite{Chen2015} --- with a large number of algorithms \emph{outperforming} humans on all the metrics. 
Finally, the metrics only work well when a number of reference translations is high \cite{Vedantam2015}, which is often unavailable, especially for current video description datasets \cite{Torabi2015}

These criticisms have led to Hodosh et al. \cite{Hodosh2013} proposing to use retrieval as a proxy for image captioning evaluation, which they argue better reflects human judgments.
Instead of generating captions, a retrieval based system ranks the available captions based on their fit to the image, and is then evaluated by assessing if the correct captions are given a high rank.
As a number of caption generation models are generative they can be used directly to assess the likelihood of a caption given an image and are being adapted by image captioning community \cite{karpathy2014deep,Kiros2014}. 
Such retrieval based evaluation metrics have also been adopted by the video captioning community \cite{Rohrbach2017}.

Visual question-answering (VQA) \cite{malinowski2015ask} task was proposed partly due to the issues facing evaluation of image captioning. VQA is a task where given an image and a question about its content the system has to answer it. Evaluating such systems is easier due to the presence of a \emph{correct} answer. However, it still faces issues such as ambiguity of certain questions and answers and question bias.

We believe that addressing the evaluation issue will be crucial for further success of multimodal translation systems. 
This will allow not only for better comparison between approaches, but also for better objectives to optimize.

\section{Alignment}
\label{sec:alignment}

\begin{table}
\begin{center}
\caption{Summary of our taxonomy for the multimodal alignment challenge. For each sub-class of our taxonomy, we include reference citations and modalities aligned.}
 \label{tab:alignment-output}
\begin{tabular}{|l|c|c|}
\hline
  \textsc{Alignment} & \textsc{Modalities} & \textsc{Reference} \\
  \hline
  \hline
  \textbf{Explicit} & &  \\
  \hline
  \hspace{0.15cm}Unsupervised & Video + Text &  \cite{Tapaswi2015a,Tapaswi2015,malmaud2015s}\\
                          & Video + Audio& \cite{zhou2009canonical, trigeorgisdeep,Noulas2012}\\
  \hline
  \hspace{0.15cm}Supervised  & Video + Text & \cite{Bojanowski2015,Zhu2015} \\
                         & Image + Text  & \cite{Plummer2015, Kong2014,Mao2016} \\
  \hline
  \hline
  \textbf{Implicit} & & \\
  \hline
  \hspace{0.15cm}Graphical models & Audio/Text + Text  & \cite{vogel1996hmm,sjolander2003hmm}\\
  \hline  
  \hspace{0.15cm}Neural networks& Image + Text & \cite{xu2015show,xiong2016dynamic, Karpathy2015} \\
   & Video + Text  & \cite{Yao2015, yu2015video}\\
  \hline
\end{tabular}
\end{center}
\end{table}

We define multimodal alignment as finding relationships and correspondences between sub-components of instances from two or more modalities. 
For example, given an image and a caption we want to find the areas of the image corresponding to the caption's words or phrases \cite{Karpathy2015}. 
Another example is, given a movie, aligning it to the script or the book chapters it was based on \cite{Zhu2015}. 

 
We categorize multimodal alignment into two types -- \emph{implicit} and \emph{explicit}. 
In explicit alignment, we are explicitly interested in aligning sub-components between modalities, e.g., aligning recipe steps with the corresponding instructional video \cite{malmaud2015s}.
Implicit alignment is used as an intermediate (often latent) step for another task, e.g., image retrieval based on text description can include an alignment step between words and image regions\cite{karpathy2014deep}.
An overview of such approaches can be seen in Table \ref{tab:alignment-output} and is presented in more detail in the following sections.



\subsection{Explicit alignment}

We categorize papers as performing explicit alignment if their main modeling objective is alignment between subcomponents of instances from two or more modalities.
A very important part of explicit alignment is the similarity metric. 
Most approaches rely on measuring similarity between sub-components in different modalities as a basic building block.
These similarities can be defined manually or learned from data.


We identify two types of algorithms that tackle explicit alignment --- \emph{unsupervised} and (weakly) \emph{supervised}. 
The first type operates with no direct alignment labels (i.e., labeled correspondences) between instances from the different modalities.
The second type has access to such (sometimes weak) labels.

\noindent \textbf{Unsupervised} multimodal alignment tackles modality alignment without requiring any direct alignment labels. 
Most of the approaches are inspired from early work on alignment for statistical machine translation \cite{Brown1993} and genome sequences \cite{Kruskal1983, Muller2007}.
To make the task easier the approaches assume certain constrains on alignment, such as temporal ordering of sequence or an existence of a similarity metric between the modalities.

Dynamic time warping (DTW) \cite{Kruskal1983, Muller2007} is a dynamic programming approach that has been extensively used to align multi-view time series. 
DTW measures the similarity between two sequences and finds an optimal match between them by time warping (inserting frames).
It requires the timesteps in the two sequences to be comparable and requires a similarity measure between them. 
DTW can be used directly for multimodal alignment  by hand-crafting similarity metrics between modalities; for example Anguera et al. \cite{anguera2014audio} use a manually defined similarity between graphemes and phonemes; and Tapaswi et al. \cite{Tapaswi2015} define a similarity between visual scenes and sentences based on appearance of same characters \cite{Tapaswi2015} to align TV shows and plot synopses.
DTW-like dynamic programming approaches have also been used for multimodal alignment of text to speech \cite{haubold2007alignment} and video \cite{Tapaswi2015a}.

As the original DTW formulation requires a pre-defined similarity metric between modalities, it was extended using canonical correlation analysis (CCA) to map the modalities to a coordinated space. This allows for both aligning (through DTW) and learning the mapping (through CCA) between different modality streams jointly and in an unsupervised manner \cite{zhou2009canonical, Shariat2011, zhou2012generalized}.
While CCA based DTW models are able to find multimodal data alignment under a linear transformation, they are not able to model non-linear relationships. 
This has been addressed by the deep canonical time warping approach \cite{trigeorgisdeep}, which can be seen as a generalization of deep CCA and DTW.


Various graphical models have also been popular for multimodal sequence alignment in an unsupervised manner.
Early work by Yu and Ballard \cite{Yu2004} used a generative graphical model to align visual objects in images with spoken words. 
A similar approach was taken by Cour et al. \cite{Cour2008} to align movie shots and scenes to the corresponding screenplay. 
Malmaud et al. \cite{malmaud2015s} used a factored HMM to align recipes to cooking videos, while Noulas et al. \cite{Noulas2012} used a dynamic Bayesian network to align speakers to videos. 
Naim et al. \cite{Naim2014} matched sentences with corresponding video frames using a hierarchical HMM model to align sentences with frames and a modified IBM  \cite{Brown1993} algorithm for word and object alignment \cite{Barnard2003}. 
This model was then extended to use latent conditional random fields for alignments \cite{Naim2015} and to incorporate verb alignment to actions in addition to nouns and objects \cite{Song2016}.


Both DTW and graphical model approaches for alignment allow for restrictions on alignment, e.g. temporal consistency, no large jumps in time, and monotonicity. 
While DTW extensions allow for learning both the similarity metric and alignment jointly, graphical model based approaches require expert knowledge for construction \cite{Yu2004,Cour2008}.

\noindent \textbf{Supervised} alignment methods rely on labeled aligned instances. 
They are used to train similarity measures that are used for aligning modalities. 

A number of supervised sequence alignment techniques take inspiration from unsupervised ones. 
Bojanowski et al. \cite{Bojanowski2014,Bojanowski2015} proposed a method similar to canonical time warping, but have also extended it to take advantage of existing (weak) supervisory alignment data for model training.
Plummer et al. \cite{Plummer2015} used CCA to find a coordinated space between image regions and phrases for alignment.
Gebru et al. \cite{gebru2016audio} trained a Gaussian mixture model and performed semi-supervised clustering together with an unsupervised latent-variable graphical model to align speakers in an audio channel with their locations in a video. 
Kong et al. \cite{Kong2014} trained a Markov random field to align objects in 3D scenes to nouns and pronouns in text descriptions.


Deep learning based approaches are becoming popular for explicit alignment (specifically for measuring similarity) due to very recent availability of aligned datasets in the language and vision communities \cite{Mao2016, Plummer2015}.  
Zhu et al. \cite{Zhu2015} aligned books with their corresponding movies/scripts by training a CNN to measure similarities between scenes and text. 
Mao et al. \cite{Mao2016} used an LSTM language model and a CNN visual one to evaluate the quality of a match between a referring expression and an object in an image. 
Yu et al. \cite{Yu2016}  extended this model to include relative appearance and context information that allows to better disambiguate between objects of the same type.
Finally, Hu et al. \cite{Hu2016} used an LSTM based scoring function to find similarities between image regions and their descriptions.

\subsection{Implicit alignment}
\label{sec:implicit-alignment}
In contrast to explicit alignment, implicit alignment is used as an intermediate (often latent) step for another task.
This allows for better performance in a number of tasks including speech recognition, machine translation, media description, and visual question-answering. 
Such models do not explicitly align data and do not rely on supervised alignment examples, but learn how to latently align the data during model training.
We identify two types of implicit alignment models: earlier work based on graphical models, and more modern neural network methods.

\noindent \textbf{Graphical models} have seen some early work used to better align words between languages for machine translation \cite{vogel1996hmm} and alignment of speech phonemes with their transcriptions \cite{sjolander2003hmm}. 
However, they require manual construction of a mapping between the modalities, for example a generative phone model that maps phonemes to acoustic features \cite{sjolander2003hmm}. 
Constructing such models requires training data or human expertise to define them manually. 

\noindent \textbf{Neural networks} 
Translation (Section \ref{sec:translation}) is an example of a modeling task that can often be improved if alignment is performed as a latent intermediate step. As we mentioned before, neural networks are popular ways to address this translation problem, using either an encoder-decoder model or through cross-modal retrieval.
When translation is performed without implicit alignment, it ends up putting a lot of weight on the encoder module to be able to properly summarize the whole image, sentence or a video with a single vectorial representation.


A very popular way to address this is through \emph{attention} \cite{Bahdanau2014}, which allows the decoder to focus on sub-components of the source instance. 
This is in contrast with encoding all source sub-components together, as is performed in a conventional encoder-decoder model. 
An attention module will tell the decoder to look more at targeted sub-components of the source to be translated --- areas of an image \cite{xu2015show}, words of a sentence \cite{Bahdanau2014}, segments of an audio sequence \cite{Chan2016,chorowski2015attention}, frames and regions in a video \cite{Yao2015,yu2015video}, and even parts of an instruction \cite{mei2015listen}.
For example, in image captioning  instead of encoding an entire image using a CNN, an attention mechanism will allow the decoder (typically an RNN) to focus on particular parts of the image when generating each successive word \cite{xu2015show}.
The attention module which learns what part of the image to focus on is typically a shallow neural network and is trained end-to-end together with a target task (e.g., translation).


Attention models have also been successfully applied to question answering tasks, as they allow for aligning the words in a question with sub-components of an information source such as a piece of text \cite{xiong2016dynamic}, an image \cite{Fukui2016}, or a video sequence \cite{Zeng2017}. 
This both allows for better performance in question answering and leads to better model interpretability \cite{Agrawal2016}.
In particular, different types of attention models have been proposed to address this problem, including hierarchical \cite{Lu2016}, stacked \cite{Yang2016}, and episodic memory attention \cite{xiong2016dynamic}.

Another neural alternative for aligning images with captions for cross-modal retrieval was proposed by Karpathy et al. \cite{karpathy2014deep, Karpathy2015}. 
Their proposed model aligns sentence fragments to image regions by using a dot product similarity measure between image region and word representations. 
While it does not use attention, it extracts a latent alignment between modalities through a similarity measure that is learned indirectly by training a retrieval model.

\subsection{Discussion}

Multimodal alignment faces a number of difficulties: 1) there are few datasets with explicitly annotated alignments; 2) it is difficult to design similarity metrics between modalities; 3) there may exist multiple possible alignments and not all elements in one modality have correspondences in another.
Earlier work on multimodal alignment focused on aligning multimodal sequences in an unsupervised manner using graphical models and dynamic programming techniques. 
It relied on hand-defined measures of similarity between the modalities or learnt them in an unsupervised manner. 
With recent availability of labeled training data supervised learning of similarities between modalities has become possible. 
However, unsupervised techniques of learning to jointly align and translate or fuse data have also become popular.



\section{Fusion}
\label{sec:fusion}
Multimodal fusion is one of the original topics in multimodal machine learning, with previous surveys emphasizing early, late and hybrid fusion approaches \cite{Zeng2009,Dmello2015}. 
In technical terms, multimodal fusion is the concept of integrating information from multiple modalities with the goal of predicting an outcome measure: a class (e.g., happy vs. sad) through classification, or a continuous value (e.g., positivity of sentiment) through regression. 
It is one of the most researched aspects of multimodal machine learning with work dating to 25 years ago \cite{Yuhas1989}. 

The interest in multimodal fusion arises from three main benefits it can provide. 
First, having access to multiple modalities that observe the same phenomenon may allow for more robust predictions. 
This has been especially explored and exploited by the AVSR community \cite{Potamianos2003}. 
Second, having access to multiple modalities might allow us to capture complementary information --- something that is not visible in individual modalities on their own. 
Third, a multimodal system can still operate when one of the modalities is missing, for example recognizing emotions from the visual signal when the person is not speaking \cite{Dmello2015}.

Multimodal fusion has a very broad range of applications, including audio-visual speech recognition (AVSR) \cite{Potamianos2003}, multimodal emotion recognition \cite{Soleymani2012}, medical image analysis \cite{James2014}, and multimedia event detection \cite{Lan2014}.
There are a number of reviews on the subject \cite{Atrey2010, Snoek2005,Zeng2009, Potamianos2003}. 
Most of them concentrate on multimodal fusion for a particular task, such as multimedia analysis, information retrieval or emotion recognition. 
In contrast, we concentrate on the machine learning approaches themselves and the technical challenges associated with these approaches. 

While some prior work used the term multimodal fusion to include all multimodal algorithms, in this survey paper we classify approaches as fusion category when the multimodal integration is performed at the later prediction stages, with the goal of predicting outcome measures. 
In recent work, the line between multimodal representation and fusion has been blurred for models such as deep neural networks where representation learning is interlaced with classification or regression objectives. 
As we will describe in this section, this line is clearer for other approaches such as graphical models and kernel-based methods.


We classify multimodal fusion into two main categories: \emph{model-agnostic} approaches (Section \ref{sec:model-free})  that are not directly dependent on a specific machine learning method; and \emph{model-based} (Section \ref{sec:model-based}) approaches that explicitly address fusion in their construction --- such as kernel-based approaches, graphical models, and neural networks.  
An overview of such approaches can be seen in Table \ref{tab:fusion}.


\begin{table}
\centering
\caption{A summary of our taxonomy of multimodal fusion approaches. \textsc{Out} --- output type (class --- classification or reg --- regression), \textsc{Temp} --- is temporal modeling possible.}
 \label{tab:fusion}
\begin{tabular}{|l|c|c|c|c|}
  \hline
  \textsc{Fusion type} & \textsc{Out} & \textsc{Temp} & \textsc{Task} & \textsc{Reference}\\
\hline
\hline
  \textbf{Model-agnostic}& & & & \\
\hline  
  \hspace{0.2cm}Early  & class & no & Emotion rec.& \cite{Castellano2008} \\
\hline
  \hspace{0.2cm}Late& reg & yes & Emotion rec.  & \cite{Ramirez2011} \\
\hline
    \hspace{0.2cm}Hybrid & class & no & MED   & \cite{Lan2014} \\
\hline
\hline
  \textbf{Model-based} & & & & \\
\hline  
  \hspace{0.15cm}Kernel-based & class & no & Object class.& \cite{Gehler2009, Bucak2014}  \\
   & class & no & Emotion rec. & \cite{Chen2014a,Jaques2015, Sikka2013} \\
\hline
  \hspace{0.15cm}Graphical   &  class & yes  & AVSR  & \cite{Gurban2008} \\
 \hspace{0.15cm}models & reg & yes  & Emotion rec. & \cite{Baltrusaitis2013}  \\
   & class & no & Media class.  & \cite{Jiang2015}  \\

\hline
  \hspace{0.15cm}Neural     & class & yes & Emotion rec. & \cite{Wollmer2010,Kahou2015}\\
\hspace{0.15cm}networks  & class & no  & AVSR  & \cite{Ngiam2011} \\
 & reg & yes  & Emotion rec.& \cite{Chen2015a}\\
\hline
\end{tabular}
\end{table}

\subsection{Model-agnostic approaches}
\label{sec:model-free}

Historically, the vast majority of multimodal fusion has been done using model-agnostic approaches \cite{Dmello2015}. Such approaches can be split into \emph{early} (i.e., feature-based), \emph{late} (i.e., decision-based) and \emph{hybrid} fusion \cite{Atrey2010}. 
Early fusion integrates features immediately after they are extracted (often by simply concatenating their representations). 
Late fusion on the other hand performs integration after each of the modalities has made a decision (e.g., classification or regression). 
Finally, hybrid fusion combines outputs from early fusion and individual unimodal predictors. 
An advantage of model agnostic approaches is that they can be implemented using almost any unimodal classifiers or regressors.

Early fusion could be seen as an initial attempt by multimodal researchers to perform multimodal representation learning --- as it can learn to exploit the correlation and interactions between low level features of each modality. 
Furthermore it only requires the training of a single model, making the training pipeline easier compared to late and hybrid fusion. 

In contrast, late fusion uses unimodal decision values and fuses them using a fusion mechanism such as averaging \cite{Shutova2016}, voting schemes \cite{Morvant2014}, weighting based on channel noise \cite{Potamianos2003} and signal variance \cite{Evangelopoulos2013}, or a learned model \cite{Glodek2011, Ramirez2011}. 
It allows for the use of different models for each modality as different predictors can model each individual modality better, allowing for more flexibility. 
Furthermore, it makes it easier to make predictions when one or more of the modalities is missing and even allows for training when no parallel data is available. 
However, late fusion ignores the low level interaction between the modalities.

Hybrid fusion attempts to exploit the advantages of both of the above described methods in a common framework. 
It has been used successfully for multimodal speaker identification \cite{Wu2005} and multimedia event detection (MED) \cite{Lan2014}.



\subsection{Model-based approaches}
\label{sec:model-based}

While model-agnostic approaches are easy to implement using unimodal machine learning methods, they end up using techniques that are not designed to cope with multimodal data. 
In this section we describe three categories of approaches that are designed to perform multimodal fusion: kernel-based methods, graphical models, and neural networks.

\noindent \textbf{Multiple kernel learning  (MKL)} methods are an extension to kernel support vector machines (SVM) that  allow for the use of different kernels for different modalities/views of the data \cite{Gonen2011}. 
As kernels can be seen as similarity functions between data points, modality-specific kernels in MKL allows for better fusion of heterogeneous data. 

MKL approaches have been an especially popular method for fusing visual descriptors for object detection \cite{Gehler2009, Bucak2014} and only recently have been overtaken by deep learning methods for the task \cite{Krizhevsky2012}. 
They have also seen use for multimodal affect recognition \cite{Chen2014a,Jaques2015, Sikka2013}, multimodal sentiment analysis \cite{Poria2015}, and multimedia event detection (MED) \cite{Yeh2012}. 
Furthermore, McFee and Lanckriet \cite{McFee2011} proposed to use MKL to perform musical artist similarity ranking from acoustic, semantic and social view data. 
Finally, Liu et al. \cite{Liu2014} used MKL for multimodal fusion in Alzheimer's disease classification. 
Their broad applicability demonstrates the strength of such approaches in various domains and across different modalities.

Besides flexibility in kernel selection, an advantage of MKL is the fact that the loss function is convex, allowing for model training using standard optimization packages and global optimum solutions \cite{Gonen2011}. 
Furthermore, MKL can be used to both perform regression and classification. 
One of the main disadvantages of MKL is the reliance on training data (support vectors) during test time, leading to slow inference and a large memory footprint. 



\noindent \textbf{Graphical models} are another family of popular methods for multimodal fusion. 
In this section we overview work done on multimodal fusion using \emph{shallow} graphical models. 
A description of deep graphical models such as deep belief networks can be found in Section \ref{sec:joint_rep}. 

Majority of graphical models can be classified into two main categories: generative --- modeling joint probability; or discriminative --- modeling conditional probability \cite{Sutton2006}. 
Some of the earliest approaches to use graphical models for multimodal fusion include generative models such as coupled \cite{Nefian2002} and factorial hidden Markov models \cite{Ghahramani1997} alongside dynamic Bayesian networks \cite{Garg2003}. 
A more recently-proposed multi-stream HMM method proposes dynamic weighting of modalities for AVSR \cite{Gurban2008}. 

Arguably, generative models lost popularity to discriminative ones such as conditional random fields (CRF) \cite{Lafferty2001} which sacrifice the modeling of joint probability for predictive power. 
A CRF model was used to better segment images by combining visual and textual information of image description \cite{Fidler2013}. 
CRF models have been extended to model latent states using hidden conditional random fields \cite{Quattoni2007} and have been applied to multimodal meeting segmentation \cite{Reiter2007}. 
Other multimodal uses of latent variable discriminative graphical models include multi-view hidden CRF \cite{Song2012a} and latent variable models \cite{Song2012}.  
More recently Jiang et al. \cite{Jiang2015} have shown the benefits of multimodal hidden conditional random fields for the task of multimedia classification.
While most graphical models are aimed at classification, CRF models have been extended to a continuous version for regression \cite{Qin2008} and applied in multimodal settings \cite{Baltrusaitis2013} for audio visual emotion recognition. 

The benefit of graphical models is their ability to easily exploit spatial and temporal structure of the data, making them especially popular for temporal modeling tasks, such as AVSR and multimodal affect recognition. 
They also allow to build in human expert knowledge into the models. and often lead to interpretable models. 






\noindent\textbf{Neural Networks} have been  used extensively for the task of multimodal fusion \cite{Ngiam2011}.
The earliest examples of using neural networks for multi-modal fusion come from work on AVSR \cite{Potamianos2003}. 
Nowadays they are being used to fuse information for visual and media question answering \cite{malinowski2015ask, xu2015ask,gao2015you}, 
gesture recognition \cite{Neverova2016}, affect analysis \cite{Kahou2015, Nojavanasghari2016}, and video description generation \cite{Jin2016}. 
While the modalities used, architectures, and optimization techniques might differ, the general idea of fusing information in joint hidden layer of a neural network remains the same.  

Neural networks have also been used for fusing temporal multimodal information through the use of RNNs and LSTMs. 
One of the earlier such applications used a bidirectional LSTM was used to perform audio-visual emotion classification \cite{Wollmer2010}. 
More recently, W\"{o}llmer et al. \cite{Wollmer2013} used LSTM models for continuous multimodal emotion recognition, demonstrating its advantage over graphical models and SVMs. Similarly, Nicolaou et al. \cite{Nicolaou2011} used LSTMs for continuous emotion prediction. 
Their proposed method used an LSTM to fuse the results from a modality specific (audio and facial expression) LSTMs.

Approaching modality fusion through recurrent neural networks has been used in various image captioning tasks, example models include: neural image captioning \cite{Vinyals2014} where a CNN image representation is decoded using an LSTM language model, gLSTM \cite{Jia2015} which incorporates the image data together with sentence decoding at every time step fusing the visual and sentence data in a joint representation. 
A more recent example is the multi-view LSTM (MV-LSTM) model proposed by Rajagopalan et al. \cite{Rajagopalan2016}. 
MV-LSTM model allows for flexible fusion of modalities in the LSTM framework by explicitly modeling the modality-specific and cross-modality interactions over time. 

A big advantage of deep neural network approaches in data fusion is their capacity to learn from large amount of data.
Secondly, recent neural architectures allow for end-to-end training of both the multimodal representation component and the fusion component. 
Finally, they show good performance when compared to non neural network based system and are able to learn complex decision boundaries that other approaches struggle with. 

The major disadvantage of neural network approaches is their lack of interpretability. 
It is difficult to tell what the prediction relies on, and which modalities or features play an important role. 
Furthermore, neural networks require large training datasets to be successful.

\subsection{Discussion}

Multimodal fusion has been a widely researched topic with a large number of approaches proposed to tackle it, including model agnostic methods, graphical models, multiple kernel learning, and various types of neural networks. 
Each approach has its own strengths and weaknesses, with some more suited for smaller datasets and others performing better in noisy environments. 
Most recently, neural networks have become a very popular way to tackle multimodal fusion, however graphical models and multiple kernel learning are still being used, especially in tasks with limited training data or where model interpretability is important.

Despite these advances multimodal fusion still faces the following challenges: 1) signals might not be temporally aligned (possibly dense continuous signal and a sparse event); 2) it is difficult to build models that exploit supplementary and not only complementary information; 3) each modality might exhibit different types and different levels of noise at different points in time.



\section{Co-learning}
\label{sec:colearning}

\begin{figure}[t]
\centering
\subfloat[Parallel \label{fig:parallel}]{\includegraphics[height=0.34\linewidth]{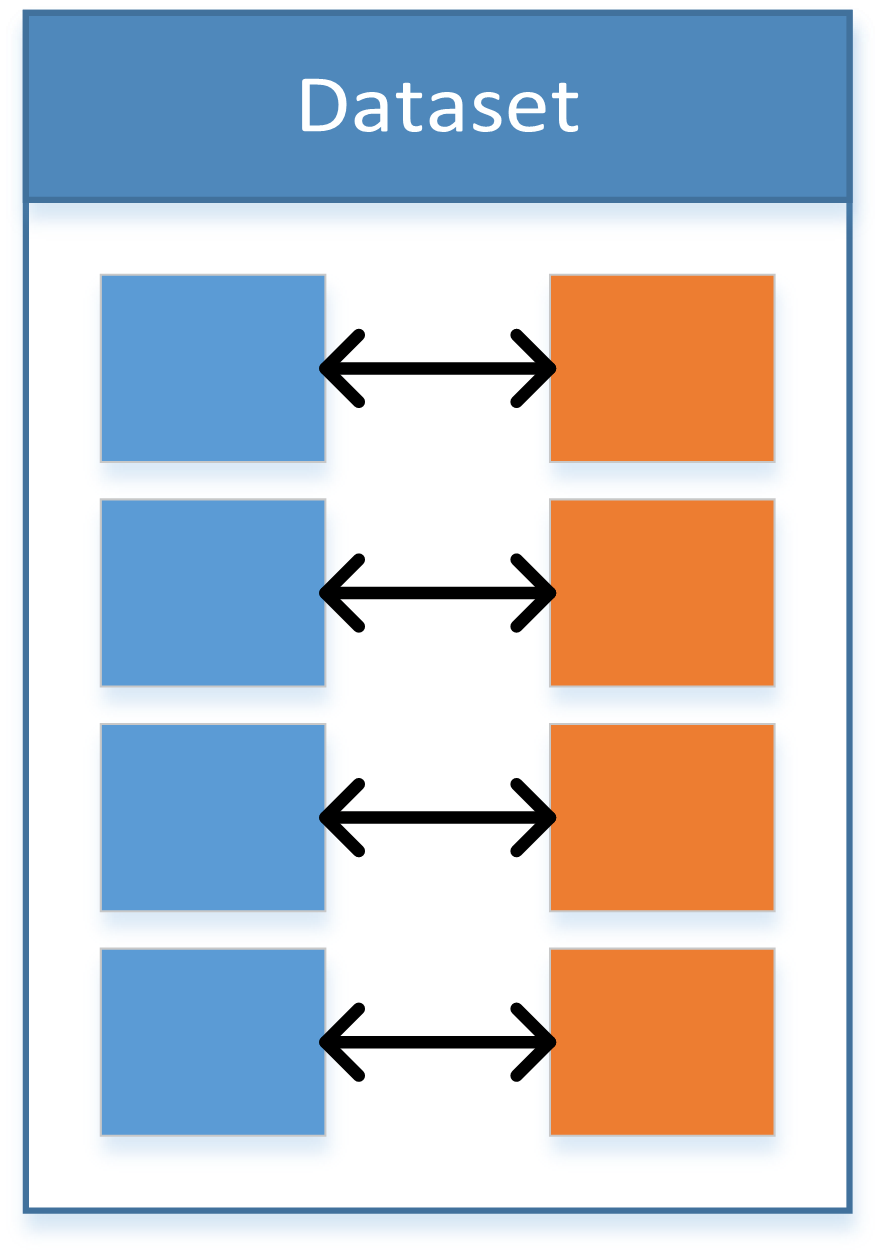}}
\hspace{0.1cm}
\subfloat[Non-parallel \label{fig:non-parallel}]{\includegraphics[height=0.34\linewidth]{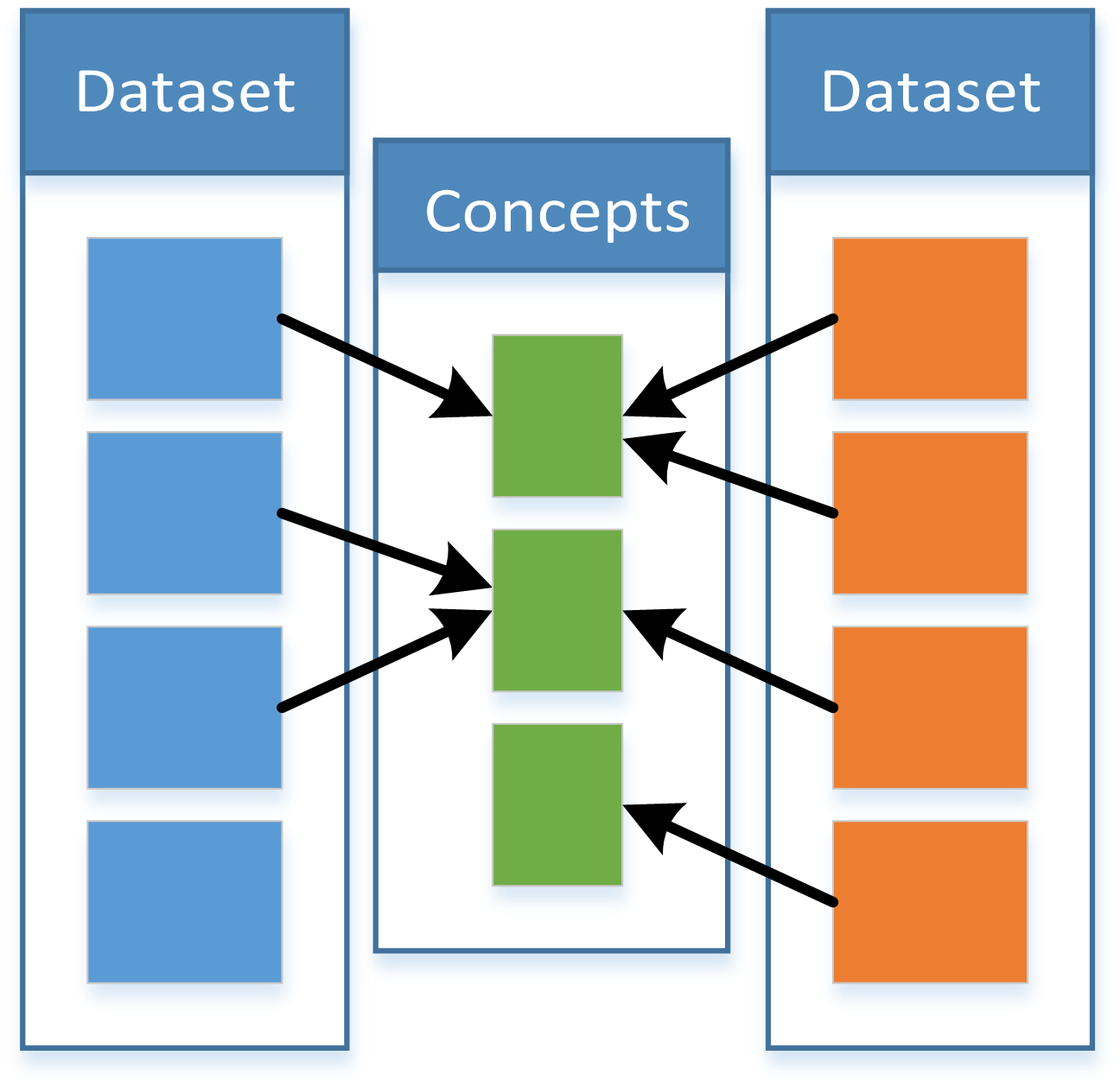}}
\hspace{0.1cm}
\subfloat[Hybrid \label{fig:bridge}]{\includegraphics[height=0.34\linewidth]{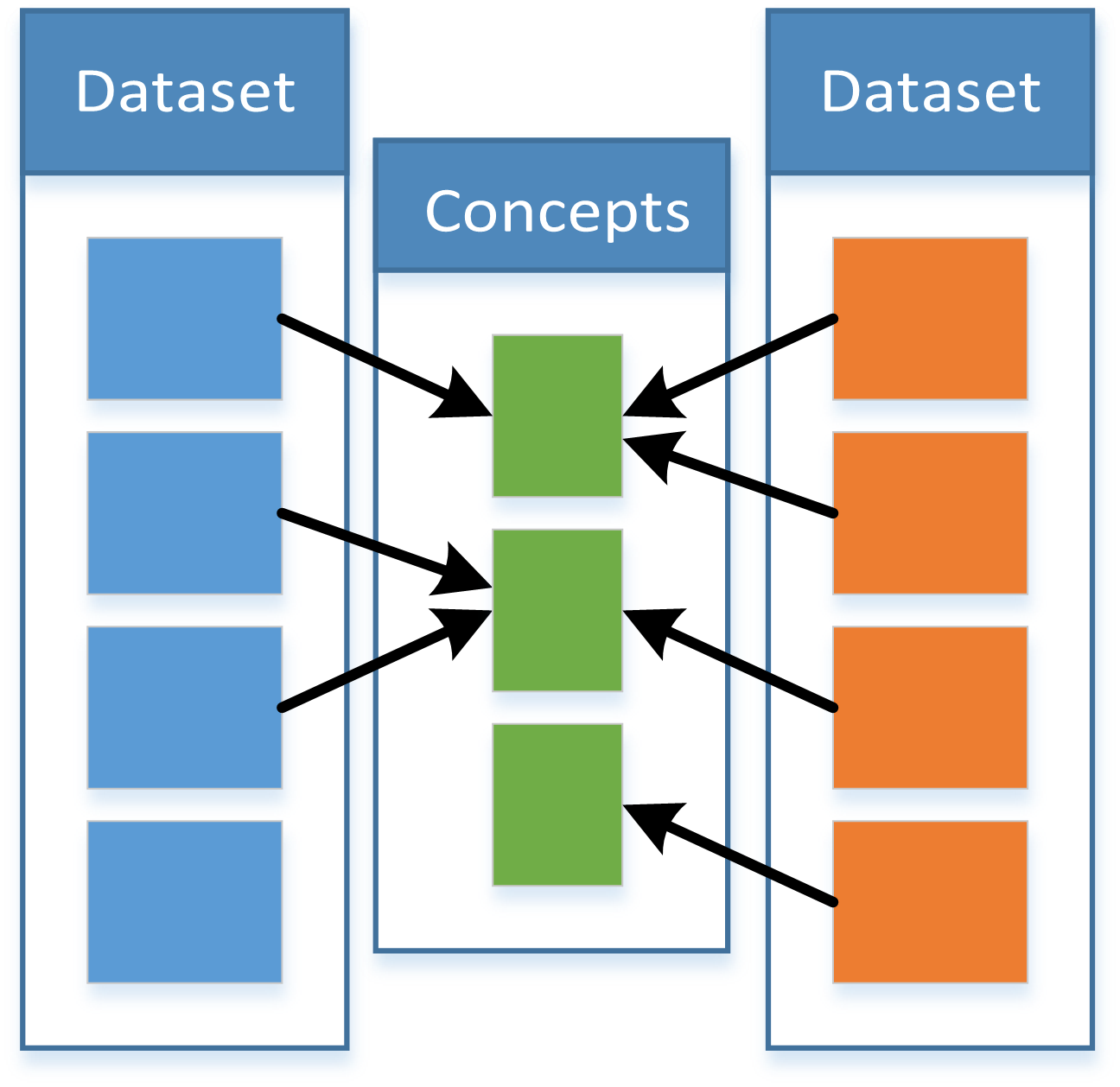}}
\caption{\label{fig:co-learning} Types of data parallelism used in co-learning: \emph{parallel} --- modalities are from the same dataset and there is a direct correspondence between instances; \emph{non-parallel} --- modalities are from different datasets and do not have overlapping instances, but overlap in general categories or concepts; \emph{hybrid} --- the instances or concepts are bridged by a third modality or a dataset. }
\end{figure}

The final multimodal challenge in our taxonomy is co-learning --- aiding the modeling of a (resource poor) modality by exploiting knowledge from another (resource rich) modality.
 It is particularly relevant when one of the modalities has limited resources --- lack of annotated data, noisy input, and unreliable labels. 
We call this challenge co-learning as most often the helper modality is used only during model training and is not used during test time.
We identify three types of co-learning approaches based on their training resources: parallel, non-parallel, and hybrid. 
\emph{Parallel-data} approaches require  training datasets where the observations from one modality are directly linked to the observations from other modalities. 
In other words, when the multimodal observations are from the same instances, such as in an audio-visual speech dataset where the video and speech samples are from the same speaker.
In contrast, \emph{non-parallel data} approaches do not require direct links between observations from different modalities. 
These approaches usually achieve co-learning by using overlap in terms of categories. 
For example, in zero shot learning when the conventional visual object recognition dataset is expanded with a second text-only dataset from Wikipedia to improve the generalization of visual object recognition.
In the \emph{hybrid} data setting the modalities are \emph{bridged} through a shared modality or a dataset.
An overview of methods in co-learning can be seen in Table \ref{tab:co-learning} and summary of data parallelism in Figure~\ref{fig:co-learning}. 

\subsection{Parallel data}
In parallel data co-learning both modalities share a set of instances --- audio recordings with the corresponding videos, images and their sentence descriptions. 
This allows for two types of algorithms to exploit that data to better model the modalities: co-training and representation learning.

\noindent\textbf{Co-training} is the process of creating more labeled training samples when we have few labeled samples in a multimodal problem \cite{Blum1998}. 
The basic algorithm builds weak classifiers in each modality to bootstrap each other with labels for the unlabeled data. 
It has been shown to discover more training samples for web-page classification based on the web-page itself and hyper-links leading in the seminal work of Blum and Mitchell \cite{Blum1998}. 
By definition this task requires parallel data as it relies on the overlap of multimodal samples. 

Co-training has been used for statistical parsing \cite{Sarkar2001} to build better visual detectors \cite{Levin2003} and for audio-visual speech recognition \cite{Christoudias2006}.
It has also been extended to deal with disagreement between modalities, by filtering out unreliable samples \cite{Christoudias2008}. 
While co-training is a powerful method for generating more labeled data, it can also lead to biased training samples resulting in overfitting.


\noindent\textbf{Transfer learning} is another way to exploit co-learning with parallel data. 
Multimodal representation learning (Section \ref{sec:joint_rep}) approaches such as multimodal deep Boltzmann machines \cite{srivastava2012multimodal} and multimodal autoencoders \cite{Ngiam2011} transfer information from representation of one modality to that of another. 
This not only leads to multimodal representations, but also to better unimodal ones, with only one modality being used during test time \cite{Ngiam2011} . 

Moon et al. \cite{Moon2015} show how to transfer information from a speech recognition neural network (based on audio) to a lip-reading one (based on images), leading to a better visual representation, and a model that can be used for lip-reading without need for audio information during test time. 
Similarly, Arora and Livescu \cite{Arora2013} build better acoustic features using CCA on acoustic and articulatory (location of lips, tongue and jaw) data. 
They use articulatory data only during CCA construction and use only the resulting acoustic (unimodal) representation during test time.


\subsection{Non-parallel data}
\label{sec:non-parallel}

Methods that rely on non-parallel data do not require the modalities to have shared instances, but only shared categories or concepts. 
Non-parallel co-learning approaches can help when learning representations, allow for better semantic concept understanding and even perform unseen object recognition.

\noindent\textbf{Transfer learning} is also possible on non-parallel data and allows to learn better representations through transferring information from a representation built using a data rich or clean modality to a data scarce or noisy modality. 
This type of trasnfer learning is often achieved by using coordinated multimodal representations (see Section \ref{sec:coord-rep}).
For example, Frome et al. \cite{Frome2013} used text to improve visual representations for image classification by coordinating CNN visual features with word2vec textual ones \cite{mikolov2013distributed} trained on separate large datasets.  
Visual representations trained in such a way result in more meaningful errors --- mistaking objects for ones of similar category \cite{Frome2013}.
Mahasseni and Todorovic \cite{Mahasseni2016} demonstrated how to regularize a color video based LSTM using an autoencoder LSTM trained on 3D skeleton data by enforcing similarities between their hidden states. 
Such an approach is able to improve the original LSTM and lead to state-of-the-art performance in action recognition.

\noindent\textbf{Conceptual grounding} refers to learning semantic meanings or concepts not purely based on language but also on additional modalities such as vision, sound, or even smell \cite{Baroni2016}.
While the majority of concept learning approaches are purely language-based, representations of meaning in humans are not merely a product of our linguistic exposure, but are also \emph{grounded} through our sensorimotor experience and perceptual system \cite{Barsalou2008, Louwerse2011}. 
Human semantic knowledge relies heavily on perceptual information \cite{Louwerse2011} and many concepts are grounded in the perceptual system and are not purely symbolic \cite{Barsalou2008}. 
This implies that learning semantic meaning purely from textual information might not be optimal, and motivates the use of visual or acoustic cues to ground our linguistic representations.

\begin{table}[t]
\centering
\caption{A summary of co-learning taxonomy, based on data parallelism. Parallel data --- multiple modalities can see the same instance. Non-parallel data --- unimodal instances are independent of each other. Hybrid data --- the modalities are \emph{pivoted} through a shared modality or dataset. }
 \label{tab:co-learning}
\begin{tabular}{|l|c|c|}
  \hline
  \textsc{Data parallelism}& \textsc{Task}& \textsc{Reference}  \\
  \hline
  \hline
  \textbf{Parallel} & & \\
\hline
 \hspace{0.3cm}  Co-training & Mixture & \cite{Blum1998,Krogel2004}\\
\hline
 \hspace{0.3cm}  Transfer learning & AVSR& \cite{Ngiam2011}  \\
 & Lip reading  & \cite{Moon2015}\\
\hline
\hline
\textbf{Non-parallel} & & \\
\hline
 \hspace{0.3cm}  Transfer learning& Visual classification& \cite{Frome2013}  \\
  \hspace{0.3cm}   & Action recognition& \cite{Mahasseni2016}\\
\hline
 \hspace{0.3cm}  Concept grounding  & Metaphor class.& \cite{Shutova2016}\\
 & Word similarity& \cite{Kiela2015a} \\
\hline
 \hspace{0.3cm} Zero shot learning  & Image class. & \cite{Socher2013,Frome2013}\\
 & Thought class.  & \cite{Palatucci2009}\\
 \hline
 \hline
\textbf{Hybrid data} & & \\
\hline
 \hspace{0.3cm} Bridging& MT and image ret.  & \cite{Rajendran2016} \\
  & Transliteration & \cite{Nakov2012} \\
\hline
\end{tabular}
\end{table}

Starting from work by Feng and Lapata \cite{Feng2010}, grounding is usually performed by finding a common latent space between the representations \cite{Feng2010, Silberer2012} (in case of parallel datasets) or by learning unimodal representations separately and then concatenating them to lead to a multimodal one \cite{Regneri2013,Shutova2016, Kiela2014,Bruni2012} (in case of non-parallel data). 
Once a multimodal representation is constructed it can be used on purely linguistic tasks. 
Shutova et al. \cite{Shutova2016} and Bruni et al. \cite{Bruni2012} used grounded representations for better classification of metaphors and literal language. 
Such representations have also been useful for measuring conceptual similarity and relatedness ---  identifying how semantically or conceptually related  two words are \cite{Kiela2014,Bruni2014,Silberer2012} or actions \cite{Regneri2013}. 
Furthermore, concepts can be grounded not only using visual signals, but also acoustic ones, leading to better performance especially on words with auditory associations \cite{Kiela2015a}, or even olfactory signals \cite{Kiela2015b} for words with smell associations.  
Finally, there is a lot of overlap between multimodal alignment and conceptual grounding, as aligning visual scenes to their descriptions leads to better textual or visual representations \cite{Regneri2013, Plummer2015,Kong2014,Yu2013}.


Conceptual grounding has been found to be an effective way to improve performance on a number of tasks. 
It also shows that language and vision (or audio) are complementary sources of information and combining them in multimodal models often improves performance. 
However, one has to be careful as grounding does not always lead to better performance \cite{Kiela2015a,Kiela2015b}, and only makes sense when grounding has relevance for the task --- such as grounding using images for visually-related concepts.

\noindent\textbf{Zero shot learning (ZSL)} refers to recognizing a concept without having explicitly seen any examples of it. 
For example classifying a cat in an image without ever having seen (labeled) images of cats. 
This is an important problem to address as in a number of tasks such as visual object classification: it is prohibitively expensive to provide training examples for every imaginable object of interest.

There are two main types of ZSL --- unimodal and multimodal. 
The unimodal ZSL looks at component parts or attributes of the object, such as phonemes to recognize an unheard word or visual attributes such as color, size, and shape to predict an unseen visual class \cite{Farhadi2009}. 
The multimodal ZSL recognizes the objects in the primary modality through the help of the secondary one --- in which the object has been seen.
The multimodal version of ZSL is a problem facing non-parallel data by definition as the overlap of seen classes is different between the modalities.

Socher et al. \cite{Socher2013} map image features to a conceptual word space and are able to classify between seen and unseen concepts. 
The unseen concepts can be then assigned to a word that is close to the visual representation --- this is enabled by the semantic space being trained on a separate dataset that has seen more concepts. 
Instead of learning a mapping from visual to concept space Frome et al. \cite{Frome2013}  learn a coordinated multimodal representation between concepts and images that allows for ZSL. 
Palatucci et al. \cite{Palatucci2009} perform prediction of words people are thinking of based on functional magnetic resonance images, they show how it is possible to predict unseen words through the use of an intermediate semantic space. 
Lazaridou et al. \cite{Lazaridou2014} present a fast mapping method for ZSL by mapping extracted visual feature vectors to text-based vectors through a neural network. 

\subsection{Hybrid data} 
In the hybrid data setting two non-parallel modalities are bridged by a shared modality or a dataset (see Figure \ref{fig:bridge}). 
The most notable example is the Bridge Correlational Neural Network \cite{Rajendran2016}, which uses a pivot modality to learn coordinated multimodal representations in presence of  non-parallel data. 
For example, in the case of multilingual image captioning, the image modality would always be paired with at least one caption in any language. 
Such methods have also been used to bridge languages that might not have parallel corpora but have access to a shared pivot language, such as for machine translation \cite{Rajendran2016, Nakov2012} and document transliteration \cite{Khapra2010}.

Instead of using a separate modality for bridging, some methods rely on existence of large datasets from a similar or related task to lead to better performance in a task that only contains limited annotated data. 
Socher and Fei-Fei \cite{Socher2010} use the existence of large text corpora in order to guide image segmentation.
While Hendricks et al. \cite{Hendricks2016} use separately trained visual model and a language model to lead to a better image and video description system, for which only limited data is available.


\subsection{Discussion}

Multimodal co-learning allows for one modality to influence the training of another, exploiting the complementary information across modalities. 
It is important to note that co-learning is task independent and could be used to create better fusion, translation, and alignment models. 
This challenge is exemplified by algorithms such as co-training, multimodal representation learning, conceptual grounding, and zero shot learning (ZSL) and has found many applications in visual classification, action recognition, audio-visual speech recognition, and semantic similarity estimation. 


\section{Conclusion}
\label{sec:conclusion}
As part of this survey, we introduced a taxonomy of multimodal machine learning: representation, translation, fusion, alignment, and co-learning. 
Some of them such as fusion have been studied for a long time, but more recent interest in representation and translation have led to a large number of new multimodal algorithms and exciting multimodal applications.

We believe that our taxonomy will help to catalog future research papers and also better understand the remaining unresolved problems facing multimodal machine learning.
\bibliographystyle{IEEEtranS}
\bibliography{tadas}

\vspace{-1.0cm}
\begin{IEEEbiography}[{\includegraphics[width=1in,clip,keepaspectratio]{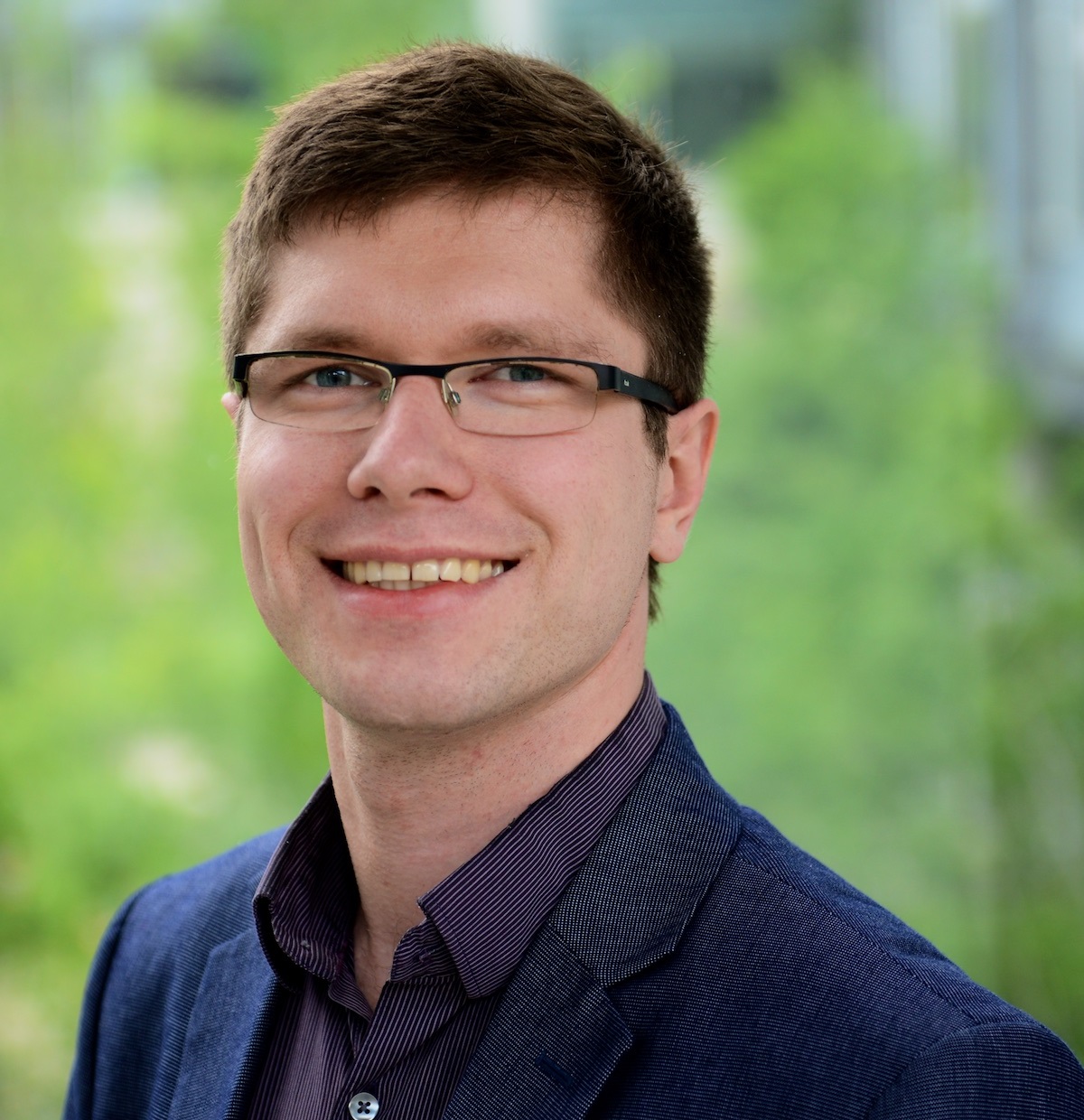}}]{Tadas Baltru\v{s}aitis}
is a post-doctoral associate at the Language Technologies Institute, Carnegie Mellon University. His primary research interests lie in the automatic understanding of non-verbal human behaviour, computer vision, and multimodal machine learning. In particular, he is interested in the application of such technologies to healthcare settings, with a particular focus on mental health. Before joining CMU, he was a post-doctoral researcher at the University of Cambridge, where he also received his Ph.D and Bachelor’s degrees in Computer Science. His Ph.D research focused on automatic facial expression analysis in especially difficult real world settings.
\end{IEEEbiography}
\vspace{-1.0cm}
\begin{IEEEbiography}[{\includegraphics[width=1in,clip,keepaspectratio]{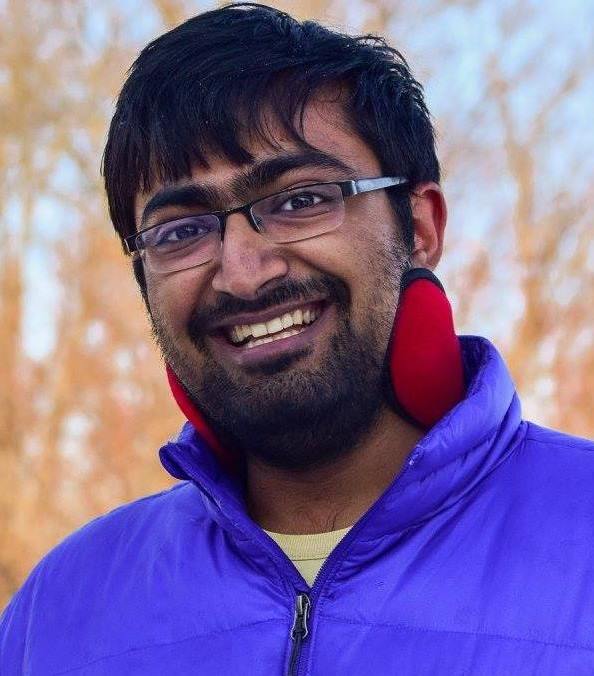}}]{Chaitanya Ahuja}
is a doctoral candidate in Language Technologies Institute in the School of Computer Science at Carnegie Mellon University. His interests range in various topics in natural language, computer vision, computational music and machine learning. Before starting with graduate school, Chaitanya completed his Bachelor's at Indian Institute of Technology, Kanpur.
\end{IEEEbiography}
\vspace{-1.0cm}
\begin{IEEEbiography}[{\includegraphics[width=1in,clip,keepaspectratio]{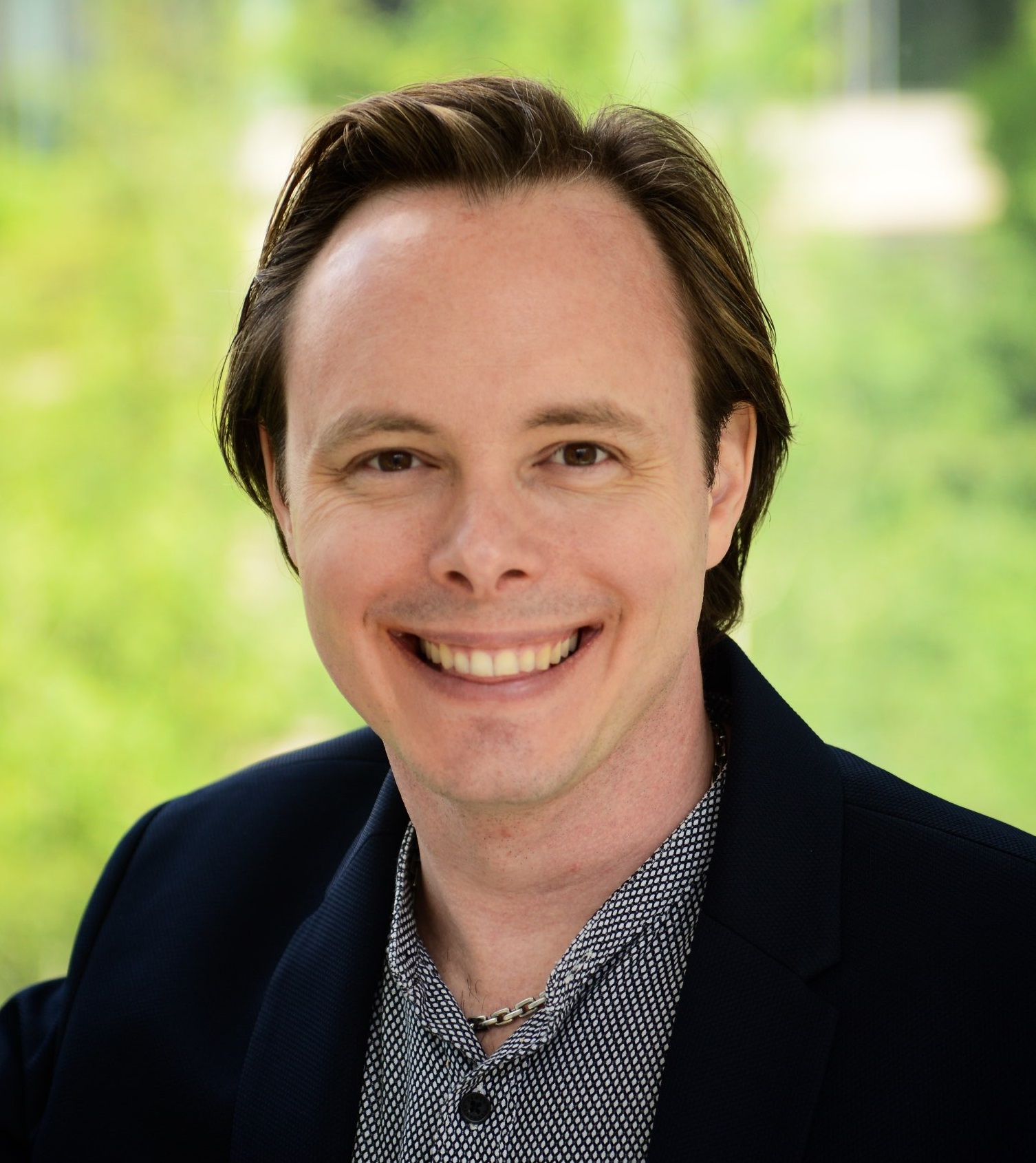}}]{Louis-Philippe Morency}
 is an Assistant Professor in the Language Technology Institute at Carnegie Mellon University where he leads the Multimodal Communication and Machine Learning Laboratory (MultiComp Lab). He was formerly research assistant professor in the Computer Sciences Department at University of Southern California and research scientist at USC Institute for Creative Technologies. Prof. Morency received his Ph.D. and Master degrees from MIT Computer Science and Artificial Intelligence Laboratory. His research focuses on building the computational foundations to enable computers with the abilities to analyze, recognize and predict subtle human communicative behaviors during social interactions. He is currently chair of the advisory committee for ACM International Conference on Multimodal Interaction and associate editor at IEEE Transactions on Affective Computing.
 \end{IEEEbiography}




\end{document}